\documentclass{article} %

\usepackage{iclr2027_conference,times}

\usepackage[utf8]{inputenc} %
\usepackage[T1]{fontenc}    %
\usepackage{hyperref}       %
\usepackage{url}            %
\usepackage{booktabs}       %
\usepackage{nicefrac}       %

\usepackage{microtype}      %
\usepackage[table,dvipsnames]{xcolor} %
\usepackage{multirow}
\usepackage[normalem]{ulem}
\usepackage{amsmath,amsfonts,bm}
\usepackage{fontawesome5}
\usepackage{twemojis}
\usepackage{comment}
\usepackage{xspace}
\usepackage{graphicx}
\usepackage{mwe}
\usepackage{subcaption}
\usepackage{pdflscape}
\usepackage{adjustbox} %
\usepackage{colortbl}
\usepackage{pifont}
\usepackage{wrapfig}
\usepackage{enumitem}
\usepackage{longtable}
\usepackage{svg}
\usepackage{makecell}
\usepackage{algorithm}
\usepackage{amssymb}
\usepackage{algpseudocode}
\usepackage{xstring}
\usepackage{tabularx}
\usepackage{siunitx}
\usepackage[super]{nth}
\usepackage[noabbrev, capitalise]{cleveref}
\usepackage[framemethod=tikz]{mdframed} %

\newcommand{\modelname}{CRONOS}

\iclrfinalcopy

\newcommand{\projecturl}{\url{https://genintel.github.io/CRONOS/}}

\title{\modelname{}: Benchmarking Counterfactual \\Physical Consistency in Video Models}

\author{%
  Le\'on Begiristain \\
  University of Freiburg\\
  Freiburg im Breisgau, Germany \\
  \texttt{begirist@cs.uni-freiburg.de} \\
  \And
  Olaf D\"unkel \\
  Max Planck Institute for Informatics \\
  Saarland Informatics Campus, Germany \\
  \texttt{oduenkel@mpi-inf.mpg.de} \\
  \AND
  Adam Kortylewski \\
  University of Freiburg\\
  Freiburg im Breisgau, Germany \\
  \texttt{kortylewski@cs.uni-freiburg.de}
}

\begin{document}

\maketitle

\lhead{Preprint}

\begin{abstract}
Video prediction is increasingly viewed as a path toward generalizable world models, yet it remains unclear whether these systems learn underlying causal structure or merely exploit superficial visual correlations for future prediction.
We introduce \modelname{}, an intervention-based benchmark to evaluate counterfactual physical consistency: whether a model's predictions of physical events respond appropriately to controlled changes in the visual input, such as variations of camera viewpoint, scene, object category, and object appearance.
Built in a photorealistic Unreal Engine environment, \modelname{} enables controlled, high-fidelity generation of videos across diverse scenes and dynamics, including sampling multiple physically plausible outcomes of the same physical event.
In contrast to previous benchmarks, %
\modelname{} systematically intervenes on four factors---camera viewpoint, scene, object category, and object appearance---while keeping the underlying physical event type, such as a fall, collision, or occlusion, fixed.
Our evaluation of recent open-source video generators reveals substantial failures in counterfactual physical consistency: prediction quality for the same physical event type is affected by appearance, environment, and particularly viewpoint changes. \modelname{} provides a controlled and reproducible testbed for diagnosing how generated video quality changes across interventions, establishing a concrete target for developing models that perform consistently across changes of multiple conditions. The dataset and evaluation code are available at \projecturl{}.

\end{abstract}

\section{Introduction}
\label{sec:intro}

Recent progress in generative video modeling has made it increasingly plausible to learn \emph{world models}---predictive models that capture how the visual world evolves over time and can support downstream reasoning and planning \citep{ha2018worldmodels}. Large-scale video diffusion models can synthesize temporally coherent, high-fidelity futures from partial observations, fueling the belief that scaling video prediction may yield generalizable predictive models of real-world dynamics \citep{ho2022videodiffusion,ho2022imagenvideo}. However, visual realism alone does not imply that these predictive systems develop \emph{causal representations} \citep{scholkopf2021toward} that capture relationships between objects, scenes, and dynamics, allowing robust predictions to remain stable under changes in viewpoint, appearance, or context. Such structured, causally meaningful representations are widely believed to be essential for robust generalization, compositional reasoning, and decision-making, as they enable models to distinguish underlying world dynamics from incidental visual correlations \citep{pearl2009causality, richens2024robust}. However, it remains unclear whether modern models acquire such representations or primarily rely on superficial statistical regularities in the data for prediction. Studying this gap requires principled evaluations that move beyond perceptual quality and directly test whether a model's predicted future responds appropriately to controlled changes in the visual input.

\begin{figure}[t]
    \centering
    \includegraphics[width=.9\linewidth]{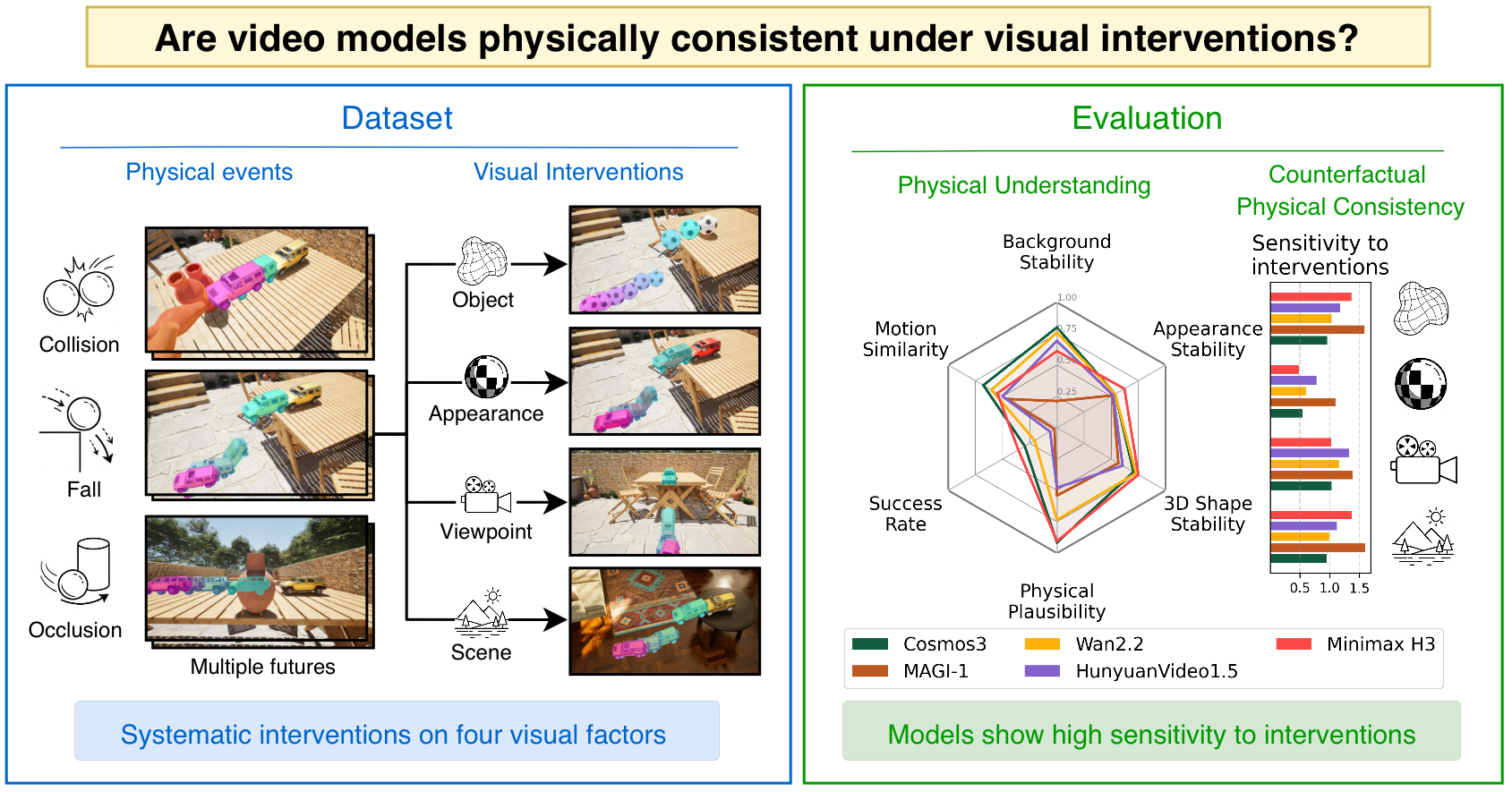}
    \caption{\textbf{The \modelname{} benchmark} evaluates
   counterfactual physical consistency: whether a model's predictions of
   physical events respond appropriately to controlled changes in the
   visual input.}
    \label{fig:teaser}
    \vspace{-15pt}
\end{figure}

Existing work has begun to probe whether video models capture physical and causal structure through specialized evaluation benchmarks. Some approaches construct controlled physics scenarios and assess predictions by comparing generated outcomes against ground-truth trajectories or physical constraints, measuring whether models obey expected dynamics such as collisions, motion, or conservation laws \citep{motamed2025generativevideomodelsunderstand, zhang2025morpheusbenchmarkingphysicalreasoning}. Other methods rely on object-centric analyses, evaluating predicted trajectories or interactions using tracking and segmentation pipelines \citep{upadhyay2026worldbench,li2025pisa}, or employ vision--language models and human judgments to detect violations of physical plausibility \citep{bansal2025videophy,ma2026out}. While these benchmarks provide insights into physical correctness, they largely evaluate predictions under a \emph{fixed visual observation}. As a result, they reveal whether a model can produce a plausible continuation of a given scene, but provide limited insight into whether the underlying predictive representation is stable and structured. A reliable model should remain stable under nuisance changes such as viewpoint or appearance variations, while adapting coherently when other aspects of the scene change. We formalize this requirement through the notion of \emph{counterfactual physical consistency}:

\begin{mdframed}[backgroundcolor=cyan!10, linecolor=gray!60!black, roundcorner=8pt, leftmargin=0pt, rightmargin=0pt, innerleftmargin=6pt, innerrightmargin=6pt]
\textbf{Counterfactual physical consistency} refers to a model's ability
to produce predictions of physical events that remain coherent across counterfactual variants of the visual input.\end{mdframed}
\vspace{-.25cm}

To study counterfactual physical consistency in modern video models, we introduce \textbf{\modelname{}}, an intervention-based benchmark designed to evaluate how predictive video models respond to controlled changes in the visual world. \modelname{} is built in a photorealistic Unreal Engine environment to enable the generation of realistic video sequences in which the underlying physical event type remains fixed while specific visual factors are systematically varied. In particular, we intervene along four complementary dimensions: camera viewpoint, scene, object category, and object appearance. Viewpoint and appearance changes primarily test robustness to nuisance variations that preserve physical parameters, while object-category and scene interventions probe whether models adapt coherently across changes in object properties and layouts.
The benchmark spans three canonical interaction scenarios---rolling and falling, collisions, and occlusion and reappearance---chosen to isolate fundamental forms of basic physical interaction. The synthetic design allows \modelname{} to also handle the challenge of physical parameter under-specification by sampling multiple physically plausible futures from the same conditioning signal. By explicitly controlling and recombining these factors, \modelname{} enables analysis of counterfactual physical consistency in video models. 

For evaluation, we introduce metrics that disentangle 3D motion from appearance, enabling a more fine-grained assessment of generation fidelity. Additionally, our intervention framework measures each model's sensitivity to controlled changes in the input signal, which serves as a diagnostic of counterfactual consistency. We apply these metrics to several state-of-the-art open-source video generation models under both image-to-video (I2V) and video-to-video (V2V) settings.

Our analyses reveal that models often fail to generate physically consistent videos and show substantial variation across intervention types, with especially high sensitivity for scene and viewpoint changes.
Further, we show that video conditioning improves over image conditioning, and that scaling model size does not necessarily lead to more consistent generation quality. The complete dataset and the evaluation code are available on our project page: \projecturl{}. An overview of the data generation and evaluation in \modelname{} can be found in \cref{fig:teaser}.

\section{Related Work}
\vspace{-0.25cm}
\textbf{Video generation models.}
Recent advances in video generation have produced models capable of synthesizing temporally coherent and visually detailed videos that are conditioned on text (T2V), images (I2V), past video frames (V2V), or combinations thereof.
Early work extended image diffusion models to the temporal domain by inserting temporal layers into latent diffusion architectures~\citep{blattmann2023align,singer2022makeavideo,blattmann2023stable,ho2022imagenvideo}.
More recently, transformer-based diffusion architectures (DiTs)~\citep{ma2024latte} have enabled models such as CogVideoX~\citep{yang2024cogvideox}, Wan~\citep{wan2025wan}, HunyuanVideo~\citep{kong2024hunyuanvideo}, and MovieGen~\citep{polyak2024moviegen} to generate high-fidelity video at scale.
Further, autoregressive formulations allow arbitrarily long generated sequences, as demonstrated by MAGI-1~\citep{teng2025magi} and Cosmos~\citep{ali2025world,agarwal2026cosmos}.
However, despite advances in terms of visual fidelity, recent studies have shown that these models frequently violate basic physical principles such as object permanence, gravity, and cause-effect relations~\citep{motamed2025generativevideomodelsunderstand,kang2024howfar,zhang2025morpheusbenchmarkingphysicalreasoning}.
This suggests that such models are limited in their ability to generalize physical understanding beyond visual patterns seen during training.
While recent efforts~\citep{li2025pisa} explored physics-aware post-training to mitigate such failures, these approaches still do not guarantee robustness.
These findings highlight a crucial gap in current video generation models that \modelname{} aims to evaluate systematically:
counterfactual physical consistency, the capability of generating videos of physical events in consistent quality even when scene parameters change.

\textbf{Evaluating video generation.}
Early evaluations of video generation models focused on image-based metrics to evaluate generation quality, such as FVD~\citep{unterthiner2018towards}, and were extended to capture various quality metrics~\citep{huang2023vbench,huang2025vbench,liu2024evalcrafter,feng2024tc}.
A growing set of benchmarks targets physical realism more directly where physical commonsense, physical laws, or scientific concepts are evaluated by humans, VLMs, or learned evaluators~\citep{bansal2024videophy,bansal2025videophy,meng2024towards,chen2025phycobench,gu2025phyworldbench,guo2025t2vphysbench,hu2025videoscience,li2025worldmodelbench,zheng2025vbench,foss2025causalvqa}.
Reference-based evaluations compare generations to trajectories, physical equations, real or simulated experiments~\citep{li2025pisa,motamed2025generativevideomodelsunderstand,zhang2025morpheusbenchmarkingphysicalreasoning,upadhyay2026worldbench,zhang2026physioneval}.
Specifically, PISA~\citep{li2025pisa} compares object trajectories of videos that cover objects in free fall scenarios.
Physics-IQ~\citep{motamed2025generativevideomodelsunderstand} evaluates videos in real-world physical experiments through image-based metrics.

In contrast, Morpheus~\citep{zhang2025morpheusbenchmarkingphysicalreasoning} measures physics-informed scores of generated videos, specifically evaluating whether equations of motion and physical constraints are satisfied.
WorldBench~\citep{upadhyay2026worldbench} estimates physical parameters of generated videos based on simple real-world physical experiments and compares the results to synthetic videos acquired from a simulation environment.
Although these reference-based evaluations move beyond perceptual metrics, each of them scores a generation against a single reference video.
The initial frames of a scene typically under-specify its outcome, so a generated video may be physically valid yet differ from the one recorded continuation and be penalized for it.
Following work on trajectory forecasting~\citep{liang2020garden}, \modelname{} instead compares each generated video against a set of physically plausible reference videos, rewarding any outcome consistent with the underlying physics rather than one particular realization of it.

Across all discussed benchmarks, evaluated sequences are independent of each other and thus measure whether a single generation is physically plausible, rather than whether a model reacts correctly when the scene producing it changes, a perspective motivated by robustness evaluations~\citep{hendrycks2019benchmarking,shu2019identifying,duenkel2025cnsbench}.
\modelname{} builds a dataset of sequences connected through controlled visual interventions, enabling the analysis of counterfactual physical consistency in video models by comparing the outcomes of different interventions.

\textbf{Simulators for probing visual understanding.}
Synthetic environments enable controlled tests that are difficult to obtain from real videos.
Many benchmarks make use of synthetic data in the realm of video reasoning:
CRAFT~\citep{ates2022craftbenchmarkcausalreasoning}, CLEVRER~\citep{yi2020clevrer}, and GRASP~\citep{jassim2024grasp} design pairs of questions and videos and evaluate models' understanding on simple scenes, while IntPhys~\citep{bordes2025intphys} focuses on detection of violations of physics.
From the modeling perspective, Physion~\citep{bear2021physion, tung2023physion++} evaluated different architectures' ability to predict the outcome of diverse physical events and PhysWorld~\citep{kang2024howfar} designed simple 2D environments to study generalization of visual properties on video diffusion.
More recently, PISA~\citep{li2025pisa} employed synthetic data to fine-tune and enhance physics modeling abilities of video models, while WorldBench~\citep{upadhyay2026worldbench} generated synthetic scenes to evaluate physical understanding.
Yet, most benchmarks leveraging synthetic data rely on objects with simple textures, and do not make use of high-fidelity rendering tools able to realistically simulate light.
In contrast, \modelname{} relies on a photorealistic simulator, keeping the advantages of a controlled environment while using higher-fidelity graphics than many synthetic physics benchmarks.

\section{\modelname{} Benchmark}
\vspace{-0.25cm}
\modelname{} frames the evaluation of video generation models as a controlled counterfactual experiment. The core experimental unit in \modelname{} is a \emph{physical event}: a basic physical simulation specified via initial states, impulses, and simulator parameters that defines a scene's 3D dynamics. From each event type, we render a set of \emph{counterfactual observations} by intervening on a single factor at a time---camera viewpoint, scene, object category, or object appearance---while holding the remaining variables fixed. Some interventions preserve the underlying physical parameters, such as viewpoint and appearance, while others change contextual or object-level properties that may alter the expected rollout. This design enables measurement of \emph{counterfactual physical consistency}: whether model predictions remain stable under nuisance interventions that do not alter the event dynamics (e.g., viewpoint) and vary coherently when interventions induce structured changes (e.g., object class).
The remainder of this section describes our controlled simulation pipeline for generating event instances (\cref{sec:method:data}), the canonical physical events defining underlying dynamics (\cref{sec:method:events}), the systematic intervention protocol used to render counterfactual observations (\cref{sec:method:interventions}), and the object-centric metrics quantifying prediction accuracy and intervention sensitivity (\cref{sec:method:metrics}).

\subsection{Data Generation}\label{sec:method:data}

\begin{figure}[t]
    \centering
\includegraphics[width=\linewidth]{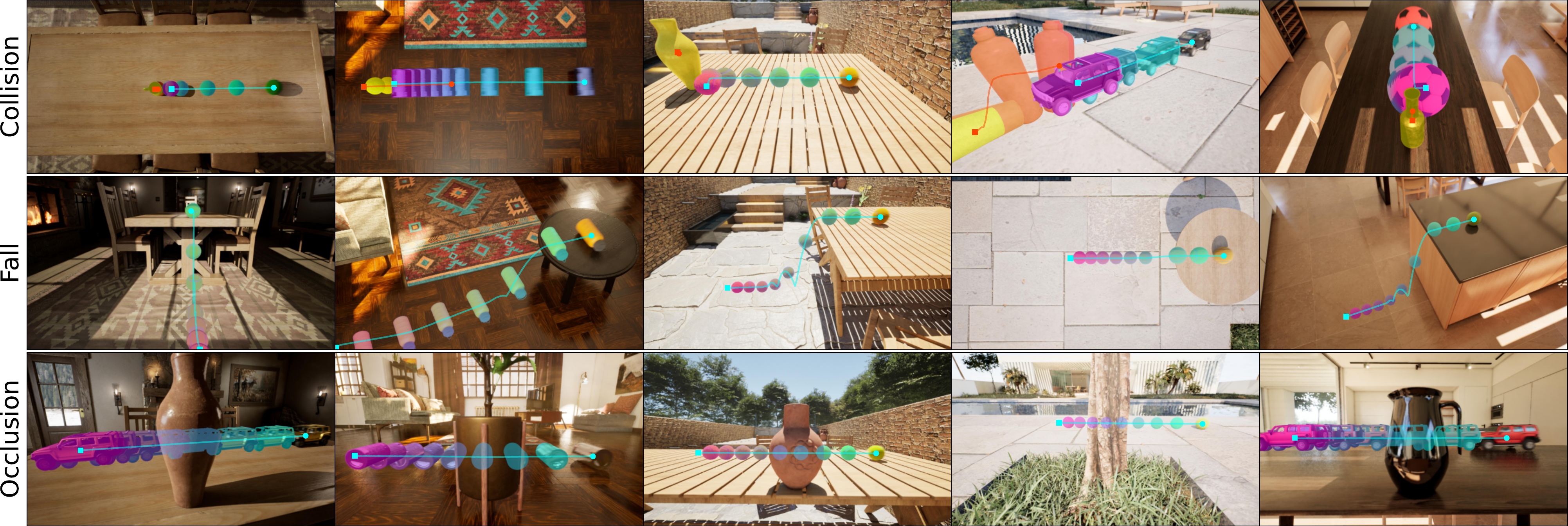}
    \caption{\textbf{\modelname{} dataset overview.}
Examples illustrating the three physical events (rows) in the \modelname{} benchmark: collision, fall, and occlusion. For each event instance, we render multiple counterfactual observations by varying factors such as scene context, camera viewpoint, object category, and object appearance. Colored overlays show object trajectories across time, visualizing the underlying motion dynamics. This controlled design enables systematic evaluation of whether video models produce consistent predictions across different observations of the same event setup.}
    \label{fig:dataset_overview}
\vspace{-15pt}
\end{figure}

We generate all sequences in a controllable Unreal Engine environment~\citep{unrealengine}.
Each event is specified by carefully selected simulator configurations, allowing targeted interventions of individual factors of realistic events.
This control is difficult to obtain from real video, where camera viewpoint, object appearance, scene context, and dynamics cannot be independently varied while preserving the same physical event type. To consider a range of physically plausible outcomes, we generate three variants of each sequence by selecting different physical parameters. All scenes are rendered at $1920\times1080$ pixels and $30$ FPS using high-quality professional 3D assets chosen to reflect common real-world environments, including indoor and outdoor environments under diverse lighting conditions.
In addition to RGB frames, the simulator provides per-object segmentation masks, used for the object-centric metrics in \cref{sec:method:metrics}.
We show examples of rendered scenes for all physical events in \cref{fig:dataset_overview}.
Additional details about the dataset can be found in \cref{sec:supplementary:dataset}.

\subsection{Physical Events}
\label{sec:method:events}

\modelname{} uses three physical events that probe complementary aspects of predictive reasoning while keeping the setup compact.
All are generated from standardized initial conditions in which an impulse initiates object motion.
Differences across intervention variants therefore come from the controlled visual change, not a new setup.
We consider three scenarios:

\textbf{Fall.} An object rolls across a surface and falls from an edge, testing prediction across changing contact conditions and free-fall motion.

\textbf{Collision.} One object impacts another, testing whether generated videos preserve physically plausible interaction dynamics, including temporal and spatial coherence and object permanence.

\textbf{Occlusion.} An object rolling across a smooth surface becomes fully occluded behind another object and later reappears, testing whether models capture long-range temporal coherence.

Together, these events provide controlled yet diverse dynamic settings for the systematic analysis of counterfactual consistency via the interventions introduced next.

\subsection{Systematic Visual Interventions}
\label{sec:method:interventions}

Building on the controlled simulation setup (\cref{sec:method:data}) and physical event dynamics (\cref{sec:method:events}), \modelname{} systematically renders a set of interventions.
For sensitivity analysis, we group variants that differ along one intervention axis while holding the remaining variables fixed:

\textbf{Scene intervention.} The background environment and scene layout are changed (e.g., height in fall sequences), testing whether models remain reliable across contextual changes and adapt to layout-dependent dynamics when scene geometry affects the rollout.

\textbf{Camera viewpoint intervention.} The rendering viewpoint is changed while keeping scene dynamics intact, probing whether models disentangle scene geometry from observed motion while maintaining perspective consistency.

\textbf{Object appearance intervention.} Visual object attributes, such as color, are changed without altering physical parameters, isolating whether models disentangle appearance from dynamics.

\textbf{Object-category intervention.} The target object is replaced with a compatible one, changing both visual properties (e.g., shape, material) and physical parameters (e.g., mass, friction), which directly affect the dynamics.
This intervention probes whether models adjust predictions coherently across object instances whose visual and physical properties differ, or instead rely on object-specific correlations learned during training.

The dataset follows a full-factorial design for each physical event, except viewpoint, which is fixed for occlusion events to preserve the intended visibility structure.
This enables fine-grained analysis of sensitivities and counterfactual consistency in generated videos.

\subsection{Evaluation Metrics}
\label{sec:method:metrics}

\modelname{} decomposes generation quality into complementary per-video metrics---background stability, motion similarity, appearance stability, 3D shape stability, and physical plausibility---together with a global success criterion that aggregates them into a single pass/fail signal per video. Physical parameter estimation from one or a few frames is inherently ill-posed, so it would be unfair to penalize models by assuming a single correct rollout. During evaluation, \modelname{} therefore compares generated videos to all available rollouts, and selects the best match. All metrics are normalized to $[0,1]$ and higher values always indicate superior performance. Additional details regarding metric computation and other steps, including segmentation masks, visibility filtering, aggregation rules, and thresholds are described in \cref{sec:supplementary:evaluation}.

\textbf{Background stability}
measures whether background regions remain coherent and fixed relative to the conditioning frame by computing pixel-wise error, following WorldBench~\citep{upadhyay2026worldbench}.
It captures artifacts such as background morphing, lighting drift, camera motion, and new objects, all undesired and explicitly excluded by the text prompt (\cref{tab:prompts}).

\textbf{Motion similarity}
compares the motion embeddings of multiple reference and the generated frames using the appearance-invariant motion encoder from DisMo~\citep{ressler2025dismo}. %

\textbf{Appearance stability}
measures whether each object preserves its visual identity over time, using cosine similarities of per-object DINOv2~\citep{oquab2023dinov2} embeddings against the initial frame, as in VBench~\citep{huang2023vbench}.
CLS tokens are computed from object-masked images. %

\textbf{3D shape stability}
measures whether object geometry changes by comparing per-frame SAM3D-reconstructed object meshes to the initial mesh via the Chamfer distance.

\textbf{Physical plausibility}
measures high-level event correctness and physical violations using a VLM-as-judge protocol~\citep{ma2026out, zheng2025vbench} with Qwen3-VL-32B~\citep{bai2025qwen3}.
This fixed set of binary questions covers common physical violations and event-completion criteria.

\textbf{Success rate} aggregates per-video metrics into a binary pass/fail criterion.
A video is successful only if all quality metrics pass their calibrated thresholds and no object disappearance is detected.
Thresholds are calibrated from the human study in \cref{sec:userstudy} by requiring equal ratios of false positive and false negative rates, where failed videos received low annotator ratings for the corresponding metric.
Further, the disappearance detector prevents segmentation failures from producing artificially high object-centric scores.
The success rate is the fraction of videos passing this test.

\textbf{Sensitivity to interventions.}
We measure how much each intervention changes the quality of generated videos. For groups of videos differing only along one intervention axis, we compute the average pairwise difference of each metric, normalize it by the average pairwise difference of that metric across seeds (shared across all models), and average across groups, seeds, and metrics. A lower sensitivity indicates higher \textbf{counterfactual physical consistency}. Note that the sensitivity measures consistency in quality rather than similarity between outputs themselves.

\section{Results}
\vspace{-0.25cm}
In the following, we present the model evaluations using our framework and discuss our findings.
After clarifying the experimental setup in \cref{sec:exp-setup}, we present the user study used to validate the selected metrics in \cref{sec:userstudy} before discussing the benchmark's findings in \cref{sec:benchmark}. Finally, we study the simulation-to-real gap with a real data experiment in \cref{sec:real_experiment}.

\subsection{Experimental Setup}\label{sec:exp-setup}

Our evaluation includes several state-of-the-art open-source video generation models:
Cosmos2.5~\citep{ali2025world}, Cosmos3-Super~\citep{agarwal2026cosmos}, CogVideoX1.5~\citep{yang2024cogvideox}, MAGI-1~\citep{teng2025magi}, Wan2.2~\citep{wan2025wan}, HunyuanVideo1.5~\citep{wu2025hunyuanvideo}, and MiniMax-H3~\citep{minimaxh3}.
For Cosmos2.5, we evaluate the two available model sizes $2$B and $14$B to study the effect of model scaling in video generation models. We run all models in the I2V setting using a single frame, and Cosmos and MAGI-1 in the V2V setting, using a five-frame clip, which reveals the initial direction and velocity while still requiring future prediction capabilities.

We use each model's standard settings and the same event-specific prompt template that describes the scene configuration, contextual details, and intended motion. Detailed text prompts can be found in \cref{sec:supplementary:prompts} and qualitative examples of generated videos are shown in \cref{supp:additional_plots}. In order to consider model variance, we sample three seeds per experiment and average the metrics across all seeds. For sensitivities, we compute the sensitivities across intervened experiments with the same seed, and then average across seeds, groups, and metrics. 

\subsection{Human Evaluation Study}\label{sec:userstudy}
\begin{figure}[t]
    \centering
    \includegraphics[width=\linewidth]{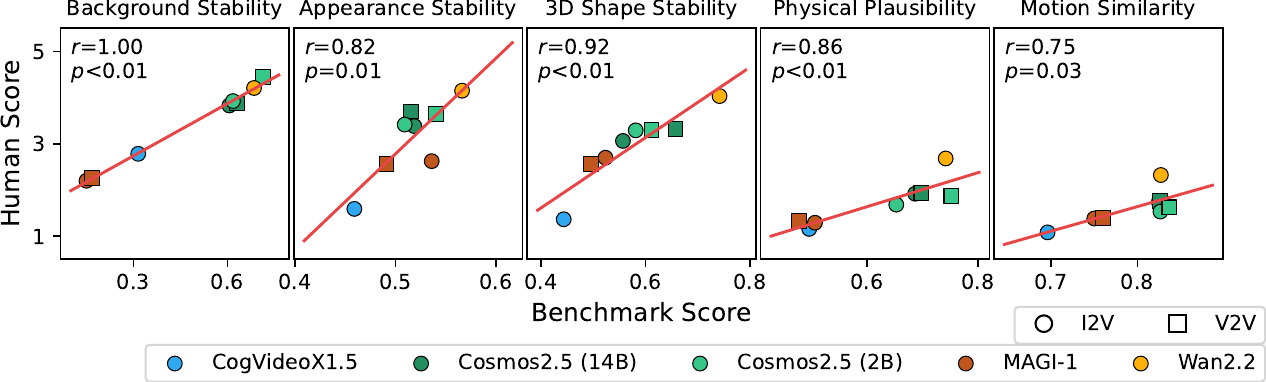}
    \caption{\textbf{Results of the human evaluation study.}
   Model performances positively correlate with human ratings (higher is better), as the Pearson correlation coefficients indicate.}
    \label{fig:user-study-results}
    \vspace{-10pt}
\end{figure}

To validate our selected metrics, we perform a user study that rates generated videos along the evaluated dimensions.
Specifically, we hire qualified annotators via Prolific and ask them to label the quality of \textit{object appearance}, \textit{object shape}, \textit{background stability}, \textit{motion plausibility}, and \textit{event quality}.
We follow \citet{bansal2025videophy} and evaluate on a scale between $1$ (very poor) and $5$ (excellent).
We select $540$ representative videos from various models with diverse objects and scenes, and we collect median-aggregated ratings of three annotators. Note that not all evaluated models are included in the study, as the most recent models were released after the study was conducted. More details including annotation instructions are provided in \cref{sec:supplementary:userstudy}.
We provide Pearson correlation coefficients for measured performance vs. human rating in \cref{fig:user-study-results}.
The stable and positive correlations support using the proposed metrics for the subsequent analysis on the complete benchmark. We also verify the stability of the success-rate thresholds in \cref{sec:supplementary:evaluation}.

\subsection{Analysis of Benchmarking Results for Video Models}\label{sec:benchmark}

\paragraph{Physical event generation.}
The averaged metrics in \cref{tab:metrics} show that performance differs substantially across models, with Cosmos3-Super and MiniMax-H3 performing best and MAGI-1 and CogVideoX1.5 being clearly worse.
Since a video is only successful if all metrics pass their calibrated thresholds, success rates are low: Cosmos3-Super (V2V) and MiniMax-H3 reach $0.30$ and $0.36$, while all other models range from $0.01$ to $0.23$.
Human ratings agree: average event quality stays below $2$ on the $1$--$5$ scale for all annotated models except Wan2.2 ($2.68$, \cref{tab:human_study}).
Failure profiles also differ across models (\cref{fig:failure_rates_per_event}): Cosmos3-Super rarely fails physical plausibility but often fails appearance stability ($45$--$55\%$ of fall and collision videos), MiniMax-H3 has the lowest appearance-stability failure rates but frequently fails background stability, and HunyuanVideo-1.5 keeps backgrounds stable but fails motion similarity in up to $88\%$ of videos. No single quality dimension therefore explains the low success rates.

\begin{mdframed}[backgroundcolor=cyan!10, linecolor=gray!60!black, roundcorner=8pt, leftmargin=0pt, rightmargin=0pt, innerleftmargin=6pt, innerrightmargin=6pt,skipbelow=0pt]
\textbf{Finding \#1}: All evaluated open-source video models struggle to reliably generate short clips of basic rigid-body physics. The strongest evaluated model achieves only $36\%$ success rate, with most models below $20\%$.\end{mdframed}
\vspace{-.5cm}
\paragraph{Visual stability versus physical correctness.}
High visual quality does not imply physical correctness. On the fall event (\cref{tab:metrics_per_event}), Wan2.2 achieves the best background stability and motion similarity of all models, yet its physical plausibility ($0.47$) falls below the calibrated threshold, which only Cosmos3-Super and MiniMax-H3 exceed. Similarly, HunyuanVideo-1.5 produces comparatively stable backgrounds but has the lowest physical plausibility overall ($0.48$). These models generate clean and stable videos of events that do not unfold as physically expected.

\begin{mdframed}[backgroundcolor=cyan!10, linecolor=gray!60!black, roundcorner=8pt, leftmargin=0pt, rightmargin=0pt, innerleftmargin=6pt, innerrightmargin=6pt]
\textbf{Finding \#2}: Visual stability does not imply physical correctness: models that generate stable backgrounds and objects can still fail to produce the expected physical event, most clearly for falling objects.\end{mdframed}
\vspace{-.5cm}
\begin{table}[t]
\centering
\small

\caption{\textbf{Benchmark performance averaged across all videos.} Metrics are normalized to $[0,1]$ and higher is better. Best, second-, and third-best values per metric are indicated by bold, underline, and dashed underline, respectively.
A video is counted as successful only if every per-video metric passes its calibrated threshold. The general low success rates indicate that meeting all quality criteria simultaneously remains challenging for current open-source video models.
}

\begin{tabular}{cccccccc}
\toprule
Mode & Model & \makecell{Bg \\ Stab.} & \makecell{Motion \\ Sim.} & \makecell{App. \\ Stab.} & \makecell{3D Shape\\ Stab.} & \makecell{Physical \\ Plaus.} & \makecell{Success \\ Rate} \\
\midrule
\multirow{4}{*}{V2V} & Cosmos2.5-2B & \underline{0.77} & \underline{0.64} & \dashuline{0.53} & 0.67 & 0.71 & \dashuline{0.23} \\
 & Cosmos2.5-14B & 0.55 & 0.56 & 0.51 & 0.64 & 0.67 & 0.14 \\
 & Cosmos3-Super & \textbf{0.80} & \textbf{0.68} & \underline{0.54} & \dashuline{0.71} & \textbf{0.92} & \underline{0.30} \\
 & Magi-1-4.5B & 0.22 & 0.47 & 0.47 & 0.58 & 0.51 & 0.01 \\
\midrule
\multirow{8}{*}{I2V} & Cosmos2.5-2B & 0.60 & 0.56 & 0.50 & 0.63 & 0.66 & 0.12 \\
 & Cosmos2.5-14B & 0.52 & 0.52 & 0.49 & 0.60 & 0.67 & 0.08 \\
 & Cosmos3-Super & 0.71 & \dashuline{0.63} & 0.50 & 0.70 & \underline{0.91} & 0.19 \\
 & Magi-1-4.5B & 0.21 & 0.46 & 0.51 & 0.57 & 0.54 & 0.02 \\
 & CogVideoX1.5-5B & 0.39 & 0.35 & 0.40 & 0.49 & 0.58 & 0.02 \\
 & Wan2.2-14B & \dashuline{0.76} & 0.60 & \underline{0.54} & \underline{0.73} & \dashuline{0.73} & 0.20 \\
 & HunyuanVideo-1.5 & 0.69 & 0.50 & 0.52 & 0.61 & 0.48 & 0.06 \\
 & MiniMax H3 & 0.61 & 0.55 & \textbf{0.62} & \textbf{0.75} & \underline{0.91} & \textbf{0.36} \\
\bottomrule
\end{tabular}
\vspace{-5pt}

\label{tab:metrics}
\end{table}

\begin{figure}[t]
    \centering
    \includegraphics[width=1\linewidth]{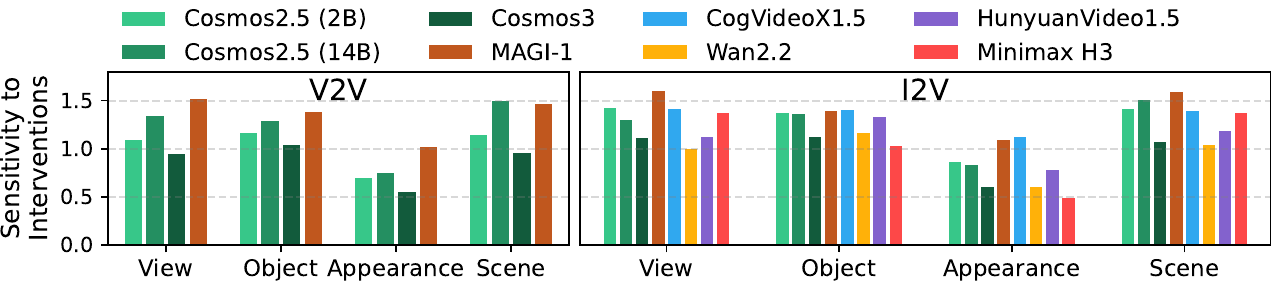}
    \caption{\textbf{Sensitivity to counterfactual interventions.}
    Sensitivities are averaged across metrics, and lower values indicate lower sensitivity.
All evaluated models show substantial variation across intervention types, including appearance changes that alter only objects' visual properties.
}
    \label{fig:sensitivities_global}
    \vspace{-10pt}
\end{figure}

\paragraph{Counterfactual consistency.}
\Cref{fig:sensitivities_global} reports each model's sensitivity to interventions. A sensitivity below $1$ indicates that an intervention changes quality less than resampling the seed, and above $1$ that it changes quality more. Since the normalization is shared across models, sensitivities are directly comparable between them.
Consistency is closely tied to benchmark performance: across the $12$ model--mode configurations, success rate and average sensitivity are strongly anti-correlated (Spearman $\rho = -0.87$), with Cosmos3-Super and Wan2.2 being the most consistent and MAGI-1 and CogVideoX1.5 the most sensitive. MiniMax-H3 is an exception: it is the most consistent model under object and appearance interventions but highly sensitive to viewpoint and scene changes, indicating a highly pixel-based world encoding.

Across models, appearance changes are tolerated best, although for MAGI-1 and CogVideoX1.5 even recoloring the object changes quality at least as much as resampling the seed (\cref{tab:sensitivities_per_event}). Scene and object interventions induce larger variation. While they change the expected rollout, a consistent model should still generate each variant with similar quality, as each video is evaluated against its own references. Viewpoint changes preserve the dynamics, object properties, and scene, yet they also induce large variation in all models. This indicates that current open-source models encode predictions in a view-dependent way, relying on visual statistics that do not transfer across viewpoints.

\begin{mdframed}[backgroundcolor=cyan!10, linecolor=gray!60!black, roundcorner=8pt, leftmargin=0pt, rightmargin=0pt, innerleftmargin=6pt, innerrightmargin=6pt]
\textbf{Finding \#3}:
Models do not achieve robust counterfactual physical consistency: generation quality changes substantially across interventions, especially for scene and viewpoint changes, and the models with the highest success rates are generally the least sensitive.
\end{mdframed}
\vspace{-.5cm}

\paragraph{Effect of video conditioning.}
Comparing I2V and V2V for Cosmos and MAGI-1 (\cref{tab:metrics}), V2V outperforms I2V in most metrics. Beyond the expected gain in motion similarity, as the conditioning frames reveal the direction and magnitude of motion, video conditioning also improves background stability for all these models and object stability in both appearance and shape for the Cosmos family. We hypothesize that conditioning frames help the model build more robust object representations, and that the static camera in the input clip is a stronger motion cue than text alone.
Video conditioning also improves consistency: average sensitivity is lower in the V2V setting for $10$ of the $12$ model--event pairs and unchanged for the other $2$ (\cref{tab:sensitivities_per_event}), as the observed motion, which is shared by all intervention variants, anchors the prediction independently of the intervened visual factors.

\begin{mdframed}[backgroundcolor=cyan!10, linecolor=gray!60!black, roundcorner=8pt, leftmargin=0pt, rightmargin=0pt, innerleftmargin=6pt, innerrightmargin=6pt]
\textbf{Finding \#4}: Video conditioning improves not only motion fidelity, as expected from the additional temporal signal, but also the stability of backgrounds and objects and the consistency of generation quality across interventions.  \end{mdframed}

\vspace{-.5cm}

\paragraph{Effect of model size.}
Comparing Cosmos2.5-2B and Cosmos2.5-14B isolates the effect of parameter count within one model family.
Surprisingly, the $14$B variant performs worse on nearly every metric in both settings (\cref{tab:metrics}), with the V2V success rate dropping from $23\%$ to $14\%$. The largest gap is in background stability, while physical plausibility is comparable and human annotators rate the motion of the $14$B variant slightly higher (\cref{tab:human_study}).
Consistency does not improve either: the $14$B variant is more sensitive than the $2$B variant on all three events in V2V and on par in I2V (\cref{tab:sensitivities_per_event}).
This agrees with \citet{kang2024howfar}, who show in a simplified 2D setting that larger video generators do not necessarily learn physical laws that generalize out of distribution, and extends this observation to photorealistic scenes and modern pretrained models.
Cosmos3-Super outperforms both Cosmos2.5 variants, but this gain cannot be attributed to scale alone, as architecture and training data also change. Since our comparison covers a single model family and scale step, it motivates a more detailed scaling study, which our benchmark enables.

\begin{mdframed}[backgroundcolor=cyan!10, linecolor=gray!60!black, roundcorner=8pt, leftmargin=0pt, rightmargin=0pt, innerleftmargin=6pt, innerrightmargin=6pt]
\textbf{Finding \#5}: Scaling Cosmos2.5 from $2$B to $14$B parameters improves neither generation quality nor consistency under interventions, indicating that model size alone does not guarantee counterfactual physical consistency. \end{mdframed}
\vspace{-.25cm}

\subsection{Real Data Experiment}
\label{sec:real_experiment}

\modelname{} makes use of photorealistic renders in order to precisely control visual factors necessary for the interventions setup. This opens the question of the simulation-to-real gap, i.e., whether findings from our simulated setup transfer to real-world data. 
We study this effect in a small-scale real-world experiment. In particular, we record a number of sequences imitating the physical events in \modelname{}'s main dataset. 
Quantitative results suggest that the gap is small, as metrics stay in the same range, consistent with the findings of \citet{upadhyay2026worldbench}. Qualitatively, we observe similar failure cases. 
We present metrics and individual rollouts in \cref{supp:real_data_experiment}.

\section{Limitations and Future Work}
\vspace{-0.25cm}
\textbf{Synthetic-to-real domain gap.}
\modelname{} uses Unreal Engine renderings rather than real videos.
This control is necessary for matched counterfactual interventions and is common in synthetic physics benchmarks~\citep{yi2020clevrer,bordes2025intphys,bear2021physion}, but it introduces a domain gap despite the high visual fidelity of the selected scenes. Experiments with real data (\cref{supp:real_data_experiment}) suggest that the failure modes and general trends are shared across data domains.

\textbf{Benchmark scope}.
\modelname{} covers three basic rigid-body interactions, chosen as canonical primitives that isolate single- and multi-object dynamics. 
Even the strongest evaluated model succeeds on only 36\% of videos. So, harder phenomena would add difficulty without making failures easier to diagnose. 
Natural extensions include further rigid-body events (ramps, stacking, containment) that reuse the current metrics unchanged, and a non-rigid track with its own evaluation.

\textbf{Scope of evaluated models.}
We evaluate open-source models with fixed weights and reproducible settings, not closed commercial systems such as Veo, Sora, or Kling. We purposefully avoid evaluating closed models to ensure scientific reproducibility and verifiability by the research community. %
On top of that, select open-source models often outperform proprietary models in common physics benchmarks~\citep{motamed2025generativevideomodelsunderstand, radsch2026physics}.

\section{Conclusions}
\vspace{-0.25cm}
We introduced \modelname{}, an intervention-based benchmark for evaluating counterfactual physical consistency in video generation models. \modelname{} consists of high-quality synthetic videos of controlled physical events, including falling dynamics, collisions, and occlusions. For each event, we systematically intervene on four visual factors---camera viewpoint, scene, object category, and object appearance---enabling fine-grained analysis of model sensitivity to structured changes.
To evaluate predictions, we combine state-of-the-art foundation models, handcrafted rules, and a VLM-as-judge protocol to design a collection of metrics focusing on object stability and physical plausibility that are validated through a user study.
Our experiments reveal substantial limitations of evaluated open-source models: even the strongest among them struggle to generate simple physical events and show substantial inconsistencies under targeted visual interventions, revealing low counterfactual physical consistency. 
Our findings suggest that many models rely on superficial visual correlations rather than producing stable predictions of scene dynamics. In general, \modelname{} provides a useful testbed for diagnosing these limitations and guiding the development of video models that produce more robust and structured predictions of the visual world.

\section*{AI disclosure}

In this work, we have used generative AI tools (coding agents) for implementing methods, in particular parts of the code on the evaluation. We have manually reviewed and tested all generated code. We have not used generative AI tools for generating synthetic datasets, developing frameworks, formulating mathematical claims, writing proofs, designing experiments, assisting with translation, formatting datasets, supporting analyses, or interpreting results. We have also used generative AI tools for formulating parts of the user survey, and drafting sections of the paper. These fragments have only been used as a starting point and manually refined. We take responsibility for the final content of this work, including text, claims, or artifacts produced with the aid of generative AI.

\section*{Ethics statement}

\textbf{Human study.} We used external annotators recruited through Prolific to gather human scores for the evaluation. Annotators participated voluntarily and were informed about the nature of the task before participation. They were compensated at a fair rate for their time and effort, in accordance with Prolific's payment guidelines. Additional information regarding the study can be found in \cref{sec:supplementary:userstudy}.

\textbf{Asset license.} 3D assets were purchased to create the dataset. We do not distribute the assets themselves, but we provide the rendered video frames and metadata. Additional details regarding the licenses and documentation can be found in \cref{sec:supplementary:asset-licenses}.

\section*{Reproducibility statement}

The evaluation code and the full dataset are publicly available on the project page (\projecturl{}). We provide additional details regarding the dataset design in \cref{sec:supplementary:dataset} and evaluation in \cref{sec:supplementary:evaluation}.

\subsubsection*{Acknowledgments}
AK acknowledges support via his Emmy Noether Research Group funded by the German Research Foundation (DFG) under grant number 468670075.
This research was funded by the Deutsche Forschungsgemeinschaft (DFG, German Research Foundation) under grant number 539134284, through EFRE (FEIH\_2698644) and the state of Baden-W\"urttemberg.
\begin{center}
\includegraphics[width=0.3\textwidth]{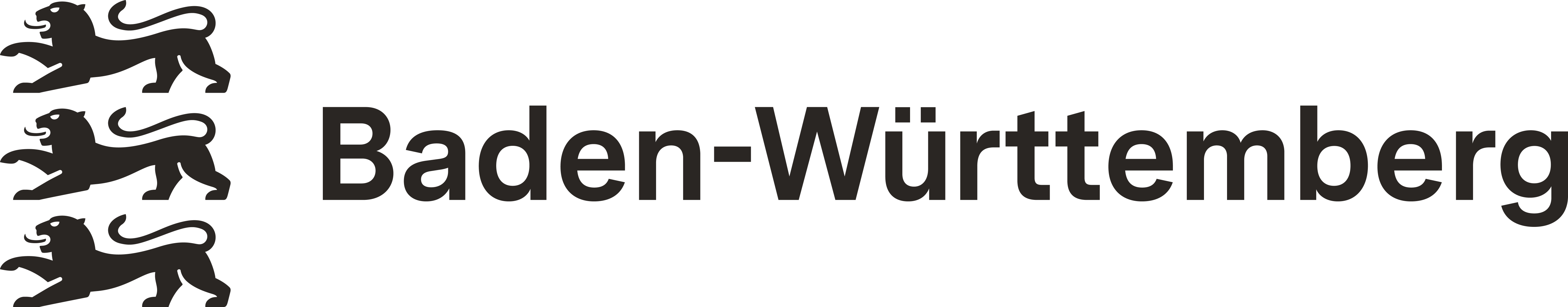} ~~~ \includegraphics[width=0.3\textwidth]{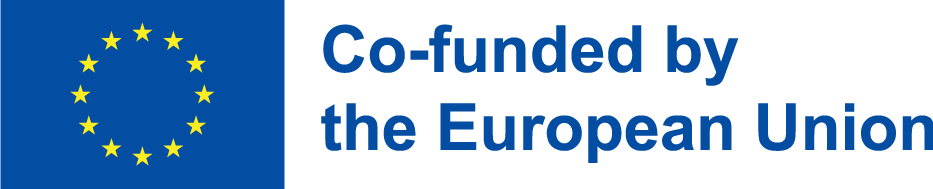}
\end{center}

\bibliographystyle{iclr2027_conference}
\bibliography{main}

\begin{thebibliography}{58}
\providecommand{\natexlab}[1]{#1}
\providecommand{\url}[1]{\texttt{#1}}
\expandafter\ifx\csname urlstyle\endcsname\relax
  \providecommand{\doi}[1]{doi: #1}\else
  \providecommand{\doi}{doi: \begingroup \urlstyle{rm}\Url}\fi

\bibitem[Agarwal et~al.(2026)Agarwal, Ali, Allen, Antolini, Aubame, Azzolini,
  Bai, Bala, Balaji, Bapst, et~al.]{agarwal2026cosmos}
Niket Agarwal, Arslan Ali, Jon Allen, Martin Antolini, Adeline Aubame, Alisson
  Azzolini, Junjie Bai, Maciej Bala, Yogesh Balaji, Josh Bapst, et~al.
\newblock {Cosmos} 3: Omnimodal world models for physical {AI}.
\newblock \emph{arXiv preprint arXiv:2606.02800}, 2026.

\bibitem[Ali et~al.(2025)Ali, Bai, Bala, Balaji, Blakeman, Cai, Cao, Cao, Cha,
  Chao, et~al.]{ali2025world}
Arslan Ali, Junjie Bai, Maciej Bala, Yogesh Balaji, Aaron Blakeman, Tiffany
  Cai, Jiaxin Cao, Tianshi Cao, Elizabeth Cha, Yu-Wei Chao, et~al.
\newblock World simulation with video foundation models for physical {AI}.
\newblock \emph{arXiv preprint arXiv:2511.00062}, 2025.

\bibitem[Ates et~al.(2022)Ates, Atesoglu, Yigit, Kesen, Kobas, Erdem, Erdem,
  Goksun, and Yuret]{ates2022craftbenchmarkcausalreasoning}
Tayfun Ates, M.~Samil Atesoglu, Cagatay Yigit, Ilker Kesen, Mert Kobas, Erkut
  Erdem, Aykut Erdem, Tilbe Goksun, and Deniz Yuret.
\newblock {CRAFT}: A benchmark for causal reasoning about forces and
  {inTeractions}, 2022.
\newblock URL \url{https://arxiv.org/abs/2012.04293}.

\bibitem[Bai et~al.(2025)Bai, Cai, Chen, Chen, Chen, Cheng, Deng, Ding, Gao,
  Ge, et~al.]{bai2025qwen3}
Shuai Bai, Yuxuan Cai, Ruizhe Chen, Keqin Chen, Xionghui Chen, Zesen Cheng,
  Lianghao Deng, Wei Ding, Chang Gao, Chunjiang Ge, et~al.
\newblock {Qwen3-VL} technical report.
\newblock \emph{arXiv preprint arXiv:2511.21631}, 2025.

\bibitem[Bansal et~al.(2024)Bansal, Lin, Xie, Zong, Yarom, Bitton, Jiang, Sun,
  Chang, and Grover]{bansal2024videophy}
Hritik Bansal, Zongyu Lin, Tianyi Xie, Zeshun Zong, Michal Yarom, Yonatan
  Bitton, Chenfanfu Jiang, Yizhou Sun, Kai-Wei Chang, and Aditya Grover.
\newblock {VideoPhy}: Evaluating physical commonsense for video generation.
\newblock \emph{arXiv preprint arXiv:2406.03520}, 2024.

\bibitem[Bansal et~al.(2025)Bansal, Peng, Bitton, Goldenberg, Grover, and
  Chang]{bansal2025videophy}
Hritik Bansal, Clark Peng, Yonatan Bitton, Roman Goldenberg, Aditya Grover, and
  Kai-Wei Chang.
\newblock {VideoPhy}-2: A challenging action-centric physical commonsense
  evaluation in video generation.
\newblock \emph{arXiv preprint arXiv:2503.06800}, 2025.

\bibitem[Bear et~al.(2021)Bear, Wang, Mrowca, Binder, Tung, Pramod, Holdaway,
  Tao, Smith, Sun, et~al.]{bear2021physion}
Daniel~M Bear, Elias Wang, Damian Mrowca, Felix~J Binder, Hsiao-Yu~Fish Tung,
  RT~Pramod, Cameron Holdaway, Sirui Tao, Kevin Smith, Fan-Yun Sun, et~al.
\newblock {Physion}: Evaluating physical prediction from vision in humans and
  machines.
\newblock \emph{arXiv preprint arXiv:2106.08261}, 2021.

\bibitem[Blattmann et~al.(2023{\natexlab{a}})Blattmann, Dockhorn, Kulal,
  Mendelevitch, Kilian, Lorenz, Levi, English, Voleti, Letts, Jampani, and
  Rombach]{blattmann2023stable}
Andreas Blattmann, Tim Dockhorn, Sumith Kulal, Daniel Mendelevitch, Maciej
  Kilian, Dominik Lorenz, Yam Levi, Zion English, Vikram Voleti, Adam Letts,
  Varun Jampani, and Robin Rombach.
\newblock {Stable Video Diffusion}: Scaling latent video diffusion models to
  large datasets.
\newblock \emph{arXiv preprint arXiv:2311.15127}, 2023{\natexlab{a}}.

\bibitem[Blattmann et~al.(2023{\natexlab{b}})Blattmann, Rombach, Ling,
  Dockhorn, Kim, Fidler, and Kreis]{blattmann2023align}
Andreas Blattmann, Robin Rombach, Huan Ling, Tim Dockhorn, Seung~Wook Kim,
  Sanja Fidler, and Karsten Kreis.
\newblock Align your latents: High-resolution video synthesis with latent
  diffusion models.
\newblock In \emph{Proceedings of the IEEE/CVF Conference on Computer Vision
  and Pattern Recognition (CVPR)}, 2023{\natexlab{b}}.

\bibitem[Bordes et~al.(2025)Bordes, Garrido, Kao, Williams, Rabbat, and
  Dupoux]{bordes2025intphys}
Florian Bordes, Quentin Garrido, Justine~T Kao, Adina Williams, Michael Rabbat,
  and Emmanuel Dupoux.
\newblock {IntPhys} 2: Benchmarking intuitive physics understanding in complex
  synthetic environments.
\newblock \emph{arXiv preprint arXiv:2506.09849}, 2025.

\bibitem[Carion et~al.(2025)Carion, Gustafson, Hu, Debnath, Hu, Suris, Ryali,
  Alwala, Khedr, Huang, et~al.]{carion2025sam}
Nicolas Carion, Laura Gustafson, Yuan-Ting Hu, Shoubhik Debnath, Ronghang Hu,
  Didac Suris, Chaitanya Ryali, Kalyan~Vasudev Alwala, Haitham Khedr, Andrew
  Huang, et~al.
\newblock {SAM} 3: Segment anything with concepts.
\newblock \emph{arXiv preprint arXiv:2511.16719}, 2025.

\bibitem[Chen et~al.(2025)Chen, Zhu, and Li]{chen2025phycobench}
Yongfan Chen, Xiuwen Zhu, and Tianyu Li.
\newblock A physical coherence benchmark for evaluating video generation models
  via optical flow-guided frame prediction, 2025.
\newblock URL \url{https://arxiv.org/abs/2502.05503}.

\bibitem[D{\"u}nkel et~al.(2025)D{\"u}nkel, Jesslen, Xie, Theobalt, Rupprecht,
  and Kortylewski]{duenkel2025cnsbench}
Olaf D{\"u}nkel, Artur Jesslen, Jiahao Xie, Christian Theobalt, Christian
  Rupprecht, and Adam Kortylewski.
\newblock {CNS-Bench}: Benchmarking image classifier robustness under
  continuous nuisance shifts.
\newblock In \emph{Proceedings of the IEEE/CVF International Conference on
  Computer Vision (ICCV)}, pp.\  19978--19988, 2025.

\bibitem[{Epic Games}(2025)]{unrealengine}
{Epic Games}.
\newblock \emph{Unreal Engine}, 2025.
\newblock URL \url{https://unrealengine.com}.
\newblock Software.

\bibitem[Feng et~al.(2024)Feng, Li, Saxon, Fu, Chen, and Wang]{feng2024tc}
Weixi Feng, Jiachen Li, Michael Saxon, Tsu-jui Fu, Wenhu Chen, and William~Yang
  Wang.
\newblock {TC-Bench}: Benchmarking temporal compositionality in text-to-video
  and image-to-video generation.
\newblock \emph{arXiv preprint arXiv:2406.08656}, 2024.

\bibitem[Foss et~al.(2025)Foss, Evans, Mitts, Sinha, Rizvi, and
  Kao]{foss2025causalvqa}
Aaron Foss, Chloe Evans, Sasha Mitts, Koustuv Sinha, Ammar Rizvi, and
  Justine~T. Kao.
\newblock {CausalVQA}: A physically grounded causal reasoning benchmark for
  video models, 2025.
\newblock URL \url{https://arxiv.org/abs/2506.09943}.

\bibitem[Gu et~al.(2025)Gu, Liu, Zeng, Nagarajan, Zhu, Hong, Fan, Yan, Zhou,
  Liu, and Wang]{gu2025phyworldbench}
Jing Gu, Xian Liu, Yu~Zeng, Ashwin Nagarajan, Fangrui Zhu, Daniel Hong, Yue
  Fan, Qianqi Yan, Kaiwen Zhou, Ming-Yu Liu, and Xin~Eric Wang.
\newblock {PhyWorldBench}: A comprehensive evaluation of physical realism in
  text-to-video models, 2025.
\newblock URL \url{https://arxiv.org/abs/2507.13428}.

\bibitem[Guo et~al.(2025)Guo, Huo, Shi, Song, Zhang, and
  Zhao]{guo2025t2vphysbench}
Xuyang Guo, Jiayan Huo, Zhenmei Shi, Zhao Song, Jiahao Zhang, and Jiale Zhao.
\newblock {T2VPhysBench}: A first-principles benchmark for physical consistency
  in text-to-video generation, 2025.
\newblock URL \url{https://arxiv.org/abs/2505.00337}.

\bibitem[Ha \& Schmidhuber(2018)Ha and Schmidhuber]{ha2018worldmodels}
David Ha and Jürgen Schmidhuber.
\newblock World models.
\newblock \emph{arXiv preprint arXiv:1803.10122}, 2018.

\bibitem[Hendrycks \& Dietterich(2019)Hendrycks and
  Dietterich]{hendrycks2019benchmarking}
Dan Hendrycks and Thomas~G. Dietterich.
\newblock Benchmarking neural network robustness to common corruptions and
  perturbations.
\newblock In \emph{7th International Conference on Learning Representations,
  {ICLR} 2019}. OpenReview.net, 2019.
\newblock URL \url{https://openreview.net/forum?id=HJz6tiCqYm}.

\bibitem[Ho et~al.(2022{\natexlab{a}})Ho, Chan, Saharia, Norouzi, and
  Fleet]{ho2022imagenvideo}
Jonathan Ho, William Chan, Chitwan Saharia, Mohammad Norouzi, and David Fleet.
\newblock {Imagen Video}: High definition video generation with diffusion
  models.
\newblock \emph{arXiv preprint arXiv:2210.02303}, 2022{\natexlab{a}}.

\bibitem[Ho et~al.(2022{\natexlab{b}})Ho, Salimans, Gritsenko, Chan, Norouzi,
  and Fleet]{ho2022videodiffusion}
Jonathan Ho, Tim Salimans, Alexey Gritsenko, William Chan, Mohammad Norouzi,
  and David~J. Fleet.
\newblock Video diffusion models.
\newblock In \emph{International Conference on Learning Representations
  (ICLR)}, 2022{\natexlab{b}}.

\bibitem[Hu et~al.(2025)Hu, Shankarampeta, Huang, Dai, Yu, Zhao, Kang, Zhao,
  Rosing, and Zhang]{hu2025videoscience}
Lanxiang Hu, Abhilash Shankarampeta, Yixin Huang, Zilin Dai, Haoyang Yu, Yujie
  Zhao, Haoqiang Kang, Daniel Zhao, Tajana Rosing, and Hao Zhang.
\newblock Benchmarking scientific understanding and reasoning for video
  generation using {VideoScience-Bench}, 2025.
\newblock URL \url{https://arxiv.org/abs/2512.02942}.

\bibitem[Huang et~al.(2025)Huang, Zhang, Xu, He, Yu, Dong, Ma, Chanpaisit, Si,
  Jiang, et~al.]{huang2025vbench}
Ziqi Huang, Fan Zhang, Xiaojie Xu, Yinan He, Jiashuo Yu, Ziyue Dong, Qianli Ma,
  Nattapol Chanpaisit, Chenyang Si, Yuming Jiang, et~al.
\newblock {VBench}++: Comprehensive and versatile benchmark suite for video
  generative models.
\newblock \emph{IEEE Transactions on Pattern Analysis and Machine
  Intelligence}, 2025.

\bibitem[Huang et~al.(2023)]{huang2023vbench}
Ziqi Huang et~al.
\newblock {VBench}: Comprehensive benchmark suite for video generative models.
\newblock \emph{arXiv preprint arXiv:2311.17982}, 2023.

\bibitem[Jassim et~al.(2024)Jassim, Holubar, Richter, Wolff, Ohmer, and
  Bruni]{jassim2024grasp}
Serwan Jassim, Mario Holubar, Annika Richter, Cornelius Wolff, Xenia Ohmer, and
  Elia Bruni.
\newblock {GRASP}: A novel benchmark for evaluating language grounding and
  situated physics understanding in multimodal language models.
\newblock In \emph{Proceedings of the Thirty-Third International Joint
  Conference on Artificial Intelligence}, pp.\  6297--6305, 2024.

\bibitem[Kang et~al.(2024)Kang, Yue, Lu, Lin, Zhao, Wang, Huang, and
  Feng]{kang2024howfar}
Bingyi Kang, Yang Yue, Rui Lu, Zhijie Lin, Yang Zhao, Kaixin Wang, Gao Huang,
  and Jiashi Feng.
\newblock How far is video generation from world model: A physical law
  perspective.
\newblock \emph{arXiv preprint arXiv:2411.02385}, 2024.

\bibitem[Karaev et~al.(2025)Karaev, Makarov, Wang, Neverova, Vedaldi, and
  Rupprecht]{karaev2025cotracker3}
Nikita Karaev, Yuri Makarov, Jianyuan Wang, Natalia Neverova, Andrea Vedaldi,
  and Christian Rupprecht.
\newblock {CoTracker3}: Simpler and better point tracking by pseudo-labelling
  real videos.
\newblock In \emph{Proceedings of the IEEE/CVF International Conference on
  Computer Vision}, pp.\  6013--6022, 2025.

\bibitem[Kong et~al.(2024)Kong, Tian, Zhang, Min, Dai, Yao, Zhu, Fang, Wu, Ai,
  et~al.]{kong2024hunyuanvideo}
Weijie Kong, Qi~Tian, Zijian Zhang, Rox Min, Zuozhuo Dai, Jin Yao, Guanyu Zhu,
  Tony Fang, Hao Wu, Yatian Ai, et~al.
\newblock {HunyuanVideo}: A systematic framework for large video generative
  models.
\newblock \emph{arXiv preprint arXiv:2412.03603}, 2024.

\bibitem[Li et~al.(2025{\natexlab{a}})Li, Michel, Pan, Liu, Roberts, and
  Xie]{li2025pisa}
Chenyu Li, Oscar Michel, Xichen Pan, Sainan Liu, Mike Roberts, and Saining Xie.
\newblock {PISA} experiments: Exploring physics post-training for video
  diffusion models by watching stuff drop.
\newblock In \emph{International Conference on Machine Learning}, pp.\
  35685--35709. PMLR, 2025{\natexlab{a}}.

\bibitem[Li et~al.(2025{\natexlab{b}})Li, Fang, Chen, Yang, Cao, Wong, Luo,
  Wang, Yin, Gonzalez, et~al.]{li2025worldmodelbench}
Dacheng Li, Yunhao Fang, Yukang Chen, Shuo Yang, Shiyi Cao, Justin Wong,
  Michael Luo, Xiaolong Wang, Hongxu Yin, Joseph~E Gonzalez, et~al.
\newblock {WorldModelBench}: Judging video generation models as world models.
\newblock \emph{arXiv preprint arXiv:2502.20694}, 2025{\natexlab{b}}.

\bibitem[Liang et~al.(2020)Liang, Jiang, Murphy, Yu, and
  Hauptmann]{liang2020garden}
Junwei Liang, Lu~Jiang, Kevin Murphy, Ting Yu, and Alexander Hauptmann.
\newblock The garden of forking paths: Towards multi-future trajectory
  prediction.
\newblock In \emph{2020 IEEE/CVF Conference on Computer Vision and Pattern
  Recognition (CVPR)}, pp.\  10505--10515. IEEE, 2020.

\bibitem[Liu et~al.(2024)Liu, Cun, Liu, Wang, Zhang, Chen, Liu, Zeng, Chan, and
  Shan]{liu2024evalcrafter}
Yaofang Liu, Xiaodong Cun, Xuebo Liu, Xintao Wang, Yong Zhang, Haoxin Chen,
  Yang Liu, Tieyong Zeng, Raymond Chan, and Ying Shan.
\newblock {EvalCrafter}: Benchmarking and evaluating large video generation
  models.
\newblock In \emph{Proceedings of the IEEE/CVF conference on computer vision
  and pattern recognition}, pp.\  22139--22149, 2024.

\bibitem[Ma et~al.(2024)Ma, Wang, Jia, Chen, Liu, Li, Chen, and
  Qiao]{ma2024latte}
Xin Ma, Yaohui Wang, Gengyun Jia, Xinyuan Chen, Ziwei Liu, Yuan-Fang Li,
  Cunjian Chen, and Yu~Qiao.
\newblock {Latte}: Latent diffusion transformer for video generation.
\newblock \emph{arXiv preprint arXiv:2401.03048}, 2024.

\bibitem[Ma et~al.(2026)Ma, Liufu, and Gkioxari]{ma2026out}
Ziqi Ma, Mengzhan Liufu, and Georgia Gkioxari.
\newblock Out of sight, out of mind? evaluating state evolution in video world
  models.
\newblock \emph{arXiv preprint arXiv:2603.13215}, 2026.

\bibitem[Meng et~al.(2024)Meng, Liao, Tan, Shao, Lu, Zhang, Cheng, Li, Qiao,
  and Luo]{meng2024towards}
Fanqing Meng, Jiaqi Liao, Xinyu Tan, Wenqi Shao, Quanfeng Lu, Kaipeng Zhang,
  Yu~Cheng, Dianqi Li, Yu~Qiao, and Ping Luo.
\newblock Towards world simulator: Crafting physical commonsense-based
  benchmark for video generation.
\newblock \emph{arXiv preprint arXiv:2410.05363}, 2024.

\bibitem[{MiniMax AI}(2026)]{minimaxh3}
{MiniMax AI}.
\newblock {MiniMax H3}: An open-weight, general-purpose, omni-modal generation
  model, 2026.
\newblock URL \url{https://www.minimax.io/blog/minimax-h3}.

\bibitem[Motamed et~al.(2026)Motamed, Culp, Swersky, Jaini, and
  Geirhos]{motamed2025generativevideomodelsunderstand}
Saman Motamed, Laura Culp, Kevin Swersky, Priyank Jaini, and Robert Geirhos.
\newblock Do generative video models understand physical principles?
\newblock In \emph{2026 IEEE/CVF Winter Conference on Applications of Computer
  Vision (WACV)}, pp.\  948--958. IEEE, 2026.

\bibitem[Oquab et~al.(2023)Oquab, Darcet, Moutakanni, Vo, Szafraniec, Khalidov,
  Fernandez, Haziza, Massa, El-Nouby, et~al.]{oquab2023dinov2}
Maxime Oquab, Timoth{\'e}e Darcet, Th{\'e}o Moutakanni, Huy Vo, Marc
  Szafraniec, Vasil Khalidov, Pierre Fernandez, Daniel Haziza, Francisco Massa,
  Alaaeldin El-Nouby, et~al.
\newblock {DINOv2}: Learning robust visual features without supervision.
\newblock \emph{arXiv preprint arXiv:2304.07193}, 2023.

\bibitem[Pearl(2009)]{pearl2009causality}
Judea Pearl.
\newblock \emph{Causality: Models, Reasoning and Inference}.
\newblock Cambridge University Press, 2009.

\bibitem[Polyak et~al.(2024)Polyak, Zohar, Brown, Tjandra, Sinha, Lee, Vyas,
  Shi, Ma, Chuang, et~al.]{polyak2024moviegen}
Adam Polyak, Amit Zohar, Andrew Brown, Andros Tjandra, Anurag Sinha, Ann Lee,
  Apoorv Vyas, Bowen Shi, Chih-Yao Ma, Ching-Yao Chuang, et~al.
\newblock {Movie Gen}: A cast of media foundation models.
\newblock \emph{arXiv preprint arXiv:2410.13720}, 2024.

\bibitem[R{\"a}dsch et~al.(2026)R{\"a}dsch, Asano, Kuehne, Bauer, Jaini,
  Geirhos, and L{\"u}th]{radsch2026physics}
Tim R{\"a}dsch, Yuki~M Asano, Hilde Kuehne, Stefan Bauer, Priyank Jaini, Robert
  Geirhos, and Carsten~T L{\"u}th.
\newblock {Physics-IQ} verified.
\newblock \emph{arXiv preprint arXiv:2606.18943}, 2026.

\bibitem[Ressler-Antal et~al.(2025)Ressler-Antal, Fundel, Alaya, Baumann,
  Krause, Gui, and Ommer]{ressler2025dismo}
Thomas Ressler-Antal, Frank Fundel, Malek~Ben Alaya, Stefan~Andreas Baumann,
  Felix Krause, Ming Gui, and Bj{\"o}rn Ommer.
\newblock Dismo: dismotangled motion representations for open-world motion
  transfer.
\newblock \emph{arXiv preprint arXiv:2511.23428}, 2025.

\bibitem[Richens \& Everitt(2024)Richens and Everitt]{richens2024robust}
Jonathan Richens and Tom Everitt.
\newblock Robust agents learn causal world models.
\newblock In \emph{The Twelfth International Conference on Learning
  Representations}, 2024.

\bibitem[Sch{\"o}lkopf et~al.(2021)Sch{\"o}lkopf, Locatello, Bauer, Ke,
  Kalchbrenner, Goyal, and Bengio]{scholkopf2021toward}
Bernhard Sch{\"o}lkopf, Francesco Locatello, Stefan Bauer, Nan~Rosemary Ke, Nal
  Kalchbrenner, Anirudh Goyal, and Yoshua Bengio.
\newblock Toward causal representation learning.
\newblock \emph{Proceedings of the IEEE}, 109\penalty0 (5):\penalty0 612--634,
  2021.

\bibitem[Shu et~al.(2020)Shu, Liu, Qiu, and Yuille]{shu2019identifying}
Michelle Shu, Chenxi Liu, Weichao Qiu, and Alan~L. Yuille.
\newblock Identifying model weakness with adversarial examiner.
\newblock In \emph{Proceedings of the AAAI Conference on Artificial
  Intelligence}, pp.\  11998--12006. {AAAI} Press, 2020.
\newblock \doi{10.1609/aaai.v34i07.6876}.

\bibitem[Singer et~al.(2022)Singer, Polyak, Hayes, Yin, An, Zhang, Hu, Yang,
  Ashual, Gafni, Parikh, Gupta, and Taigman]{singer2022makeavideo}
Uriel Singer, Adam Polyak, Thomas Hayes, Xi~Yin, Jie An, Songyang Zhang, Qiyuan
  Hu, Harry Yang, Oron Ashual, Oran Gafni, Devi Parikh, Sonal Gupta, and Yaniv
  Taigman.
\newblock {Make-A-Video}: Text-to-video generation without text-video data.
\newblock \emph{arXiv preprint arXiv:2209.14792}, 2022.

\bibitem[Teng et~al.(2025)Teng, Jia, Sun, Li, Li, Tang, Han, Zhang, Zhang, Luo,
  et~al.]{teng2025magi}
Hansi Teng, Hongyu Jia, Lei Sun, Lingzhi Li, Maolin Li, Mingqiu Tang, Shuai
  Han, Tianning Zhang, WQ~Zhang, Weifeng Luo, et~al.
\newblock {MAGI-1}: Autoregressive video generation at scale.
\newblock \emph{arXiv preprint arXiv:2505.13211}, 2025.

\bibitem[Tung et~al.(2023)Tung, Ding, Chen, Bear, Gan, Tenenbaum, Yamins, Fan,
  and Smith]{tung2023physion++}
Hsiao-Yu Tung, Mingyu Ding, Zhenfang Chen, Daniel Bear, Chuang Gan, Josh
  Tenenbaum, Dan Yamins, Judith Fan, and Kevin Smith.
\newblock {Physion}++: Evaluating physical scene understanding that requires
  online inference of different physical properties.
\newblock \emph{Advances in Neural Information Processing Systems},
  36:\penalty0 67048--67068, 2023.

\bibitem[Unterthiner et~al.(2018)Unterthiner, Van~Steenkiste, Kurach, Marinier,
  Michalski, and Gelly]{unterthiner2018towards}
Thomas Unterthiner, Sjoerd Van~Steenkiste, Karol Kurach, Raphael Marinier,
  Marcin Michalski, and Sylvain Gelly.
\newblock Towards accurate generative models of video: A new metric \&
  challenges.
\newblock \emph{arXiv preprint arXiv:1812.01717}, 2018.

\bibitem[Upadhyay et~al.(2026)Upadhyay, Zhang, Solomon, Agrawal, Boreddy,
  Narayana, Ba, Wong, de~Melo, and Kadambi]{upadhyay2026worldbench}
Rishi Upadhyay, Howard Zhang, Jim Solomon, Ayush Agrawal, Pranay Boreddy,
  Shruti~Satya Narayana, Yunhao Ba, Alex Wong, Celso~M de~Melo, and Achuta
  Kadambi.
\newblock {WorldBench}: Disambiguating physics for diagnostic evaluation of
  world models.
\newblock \emph{arXiv preprint arXiv:2601.21282}, 2026.

\bibitem[Wan et~al.(2025)Wan, Wang, Ai, Wen, Mao, Xie, Chen, Yu, Zhao, Yang,
  et~al.]{wan2025wan}
Team Wan, Ang Wang, Baole Ai, Bin Wen, Chaojie Mao, Chen-Wei Xie, Di~Chen,
  Feiwu Yu, Haiming Zhao, Jianxiao Yang, et~al.
\newblock {Wan}: Open and advanced large-scale video generative models.
\newblock \emph{arXiv preprint arXiv:2503.20314}, 2025.

\bibitem[Wu et~al.(2025)Wu, Zou, Li, Huang, Yang, Tan, Peng, Wu, Xiong, Jiang,
  et~al.]{wu2025hunyuanvideo}
Bing Wu, Chang Zou, Changlin Li, Duojun Huang, Fang Yang, Hao Tan, Jack Peng,
  Jianbing Wu, Jiangfeng Xiong, Jie Jiang, et~al.
\newblock {HunyuanVideo} 1.5 technical report.
\newblock \emph{arXiv preprint arXiv:2511.18870}, 2025.

\bibitem[Yang et~al.(2024)Yang, Teng, Zheng, Ding, Huang, Xu, Yang, Hong,
  Zhang, Feng, et~al.]{yang2024cogvideox}
Zhuoyi Yang, Jiayan Teng, Wendi Zheng, Ming Ding, Shiyu Huang, Jiazheng Xu,
  Yuanming Yang, Wenyi Hong, Xiaohan Zhang, Guanyu Feng, et~al.
\newblock {CogVideoX}: Text-to-video diffusion models with an expert
  transformer.
\newblock \emph{arXiv preprint arXiv:2408.06072}, 2024.

\bibitem[Yi et~al.(2020)Yi, Gan, Li, Kohli, Wu, Torralba, and
  Tenenbaum]{yi2020clevrer}
Kexin Yi, Chuang Gan, Yunzhu Li, Pushmeet Kohli, Jiajun Wu, Antonio Torralba,
  and Joshua~B Tenenbaum.
\newblock {CLEVRER}: {CoLlision} events for video {REpresentation} and
  reasoning.
\newblock In \emph{International Conference on Learning Representations}, 2020.

\bibitem[Zhang et~al.(2025)Zhang, Cherniavskii, Tragoudaras, Vozikis, Nijdam,
  Prinzhorn, Bodracska, Sebe, Zadaianchuk, and
  Gavves]{zhang2025morpheusbenchmarkingphysicalreasoning}
Chenyu Zhang, Daniil Cherniavskii, Antonios Tragoudaras, Antonios Vozikis,
  Thijmen Nijdam, Derck W.~E. Prinzhorn, Mark Bodracska, Nicu Sebe, Andrii
  Zadaianchuk, and Efstratios Gavves.
\newblock {Morpheus}: Benchmarking physical reasoning of video generative
  models with real physical experiments, 2025.
\newblock URL \url{https://arxiv.org/abs/2504.02918}.

\bibitem[Zhang et~al.(2026)Zhang, Jing, Yu, Ding, Nie, Wang, Du, Zou, Wu, and
  Shuai]{zhang2026physioneval}
Qin Zhang, Peiyu Jing, Hong-Xing Yu, Fangqiang Ding, Fan Nie, Weimin Wang,
  Yilun Du, James Zou, Jiajun Wu, and Bing Shuai.
\newblock {Physion-Eval}: Evaluating physical realism in generated video via
  human reasoning, 2026.
\newblock URL \url{https://arxiv.org/abs/2603.19607}.

\bibitem[Zheng et~al.(2025)Zheng, Huang, Liu, Zou, He, Zhang, Gu, Zhang, He,
  Zheng, et~al.]{zheng2025vbench}
Dian Zheng, Ziqi Huang, Hongbo Liu, Kai Zou, Yinan He, Fan Zhang, Lulu Gu,
  Yuanhan Zhang, Jingwen He, Wei-Shi Zheng, et~al.
\newblock {VBench}-2.0: Advancing video generation benchmark suite for
  intrinsic faithfulness.
\newblock \emph{arXiv preprint arXiv:2503.21755}, 2025.

\end{thebibliography}

\newpage

\appendix

 {
        \centering
        \Large
        Appendix \\
        \vspace{1.5em}
   }

\section{Dataset Details}
\label{sec:supplementary:dataset}

The dataset of \modelname{} is handcrafted using combinations of object and scene assets. Objects are selected to align with the physical events under study, ensuring that their geometry and material properties allow sliding or rolling over smooth surfaces, thereby producing visually plausible and dynamically rich motion patterns.

Simulated physical parameters, including mass, friction, and restitution coefficients, are explicitly defined per object and are kept consistent across events. In order to render multiple variations of each possible configuration, we select key physical parameters of the physical events and modify them. In particular, we select a high, medium and low value for: moving object mass (collision), elasticity (fall) and friction (occlusion). This way, we cover the range of plausible physical behaviors, while preserving the systematic interventions intact.

\begin{figure}[htbp]
    \centering
    \includegraphics[width=1.0\linewidth]{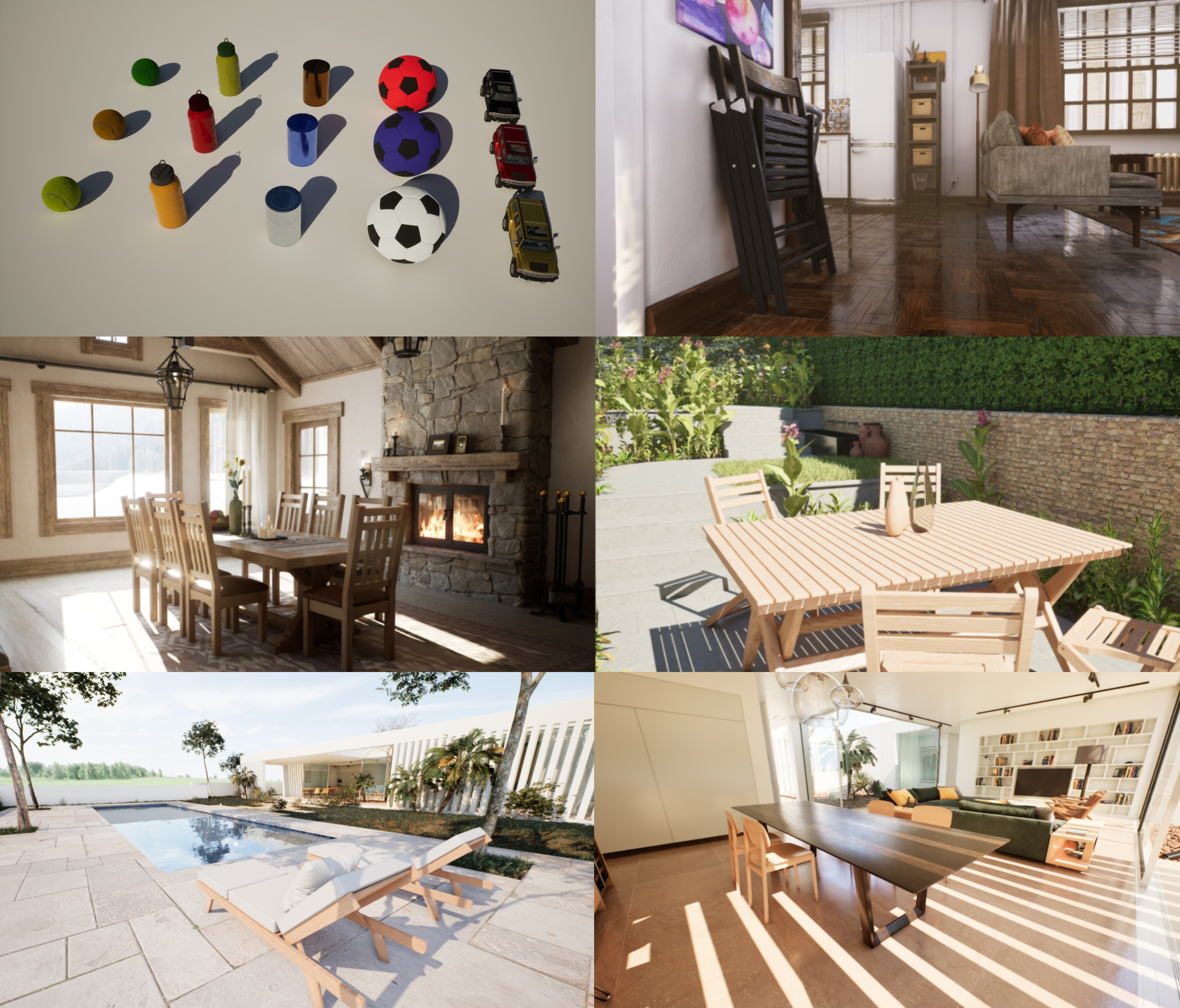}
    \caption{\textbf{Dataset assets.} Overview of Unreal Engine assets used for all rendered sequences: original scenes, object models, and appearance variations.}
    \label{fig:assets}
\end{figure}

Scene layouts and initial conditions (e.g., object positions and applied forces), are individually tailored to fit each physical event, and camera viewpoints are strategically positioned to preserve global context while maintaining clear visibility of the relevant interactions throughout the sequence. This ensures that differences between rendered videos are attributable only to controlled interventions on the scene state. The proposed controlled generation process forms the basis for the construction of matched counterfactual observations used throughout \modelname{}. \cref{tab:interventions} shows the number of different variations per intervention type. An overview of the original asset scenes, objects used for interventions and their different appearances are shown in \cref{fig:assets}. For all three physical events, all variations are rendered. The only exception is the occlusion event, where the view is not altered to preserve the intended visibility effect. The number of videos rendered per event is shown in \cref{tab:videos}.

\begin{table}[htbp]
\centering
\begin{minipage}{0.49\textwidth}
\centering
\caption{\textbf{Number of variations per intervention.}}
\begin{tabular}{lc}
\toprule
\textbf{Intervention} & \multicolumn{1}{l}{\textbf{\# of variations}} \\ \midrule
Scene       & $5$                                    \\
Object      & $5$                                    \\
Appearance  & $3$                                    \\
View        & $4$                                    \\
\bottomrule
\end{tabular}
\label{tab:interventions}
\end{minipage}
\hfill
\begin{minipage}{0.49\textwidth}
\centering
\caption{\textbf{Number of videos per physical event.}}
\begin{tabular}{lc}
\toprule
\textbf{Event}     & \multicolumn{1}{l}{\textbf{\# of videos}} \\ \midrule
Fall & $900$                              \\
Collision      & $900$                              \\
Occlusion & $225$                               \\ \midrule
Total     & $2025$                              \\
\bottomrule
\end{tabular}
\label{tab:videos}
\end{minipage}
\end{table}

\section{Text Prompts}
\label{sec:supplementary:prompts}

In addition to the visual conditioning signal, text prompts are provided to all video models. These prompts have been manually designed for each physical event, and follow the same structure across interventions, as shown in \cref{tab:prompts}. The prompts coarsely describe each physical event, serving as an overview of the expected dynamics but without providing many details. The specific variation values of each intervention type are added through the prompt, as well as details required by each physical event (e.g., the collision target and the occluding object). On top of this, an additional shared prompt is added in all cases, providing general details about the expected physical behavior and camera movement.

\begin{table}[htbp]
\centering
\caption{\textbf{Structure of text prompts.} General text prompt structure for each physical event. Variables in brackets are substituted using the details of each particular video. An additional prompt describing the general physical behavior and expected movement is added to all physical events.}
\begin{tabular}{lp{8cm}}
\toprule
\textbf{Event}             & \textbf{Prompt Structure }                                                                                                                                                                                                                                                                                                                        \\ \midrule
Fall              & The video shows a \{view\} view of a \{appearance\}\{object\} smoothly \{object\_movement\} across a \{surface\} in a \{scene\}. When the \{object\} reaches the edge, it falls off the \{surface\} and vertically descends until it hits the ground. The object maintains its shape and does not break during the fall.                 \\ \midrule
Collision         & The video shows a \{view\} view of a \{appearance\}\{object\} smoothly \{object\_movement\} across a \{surface\}, colliding with a \{collision\_target\} in a \{scene\}. The objects react to the collision but preserve their rigidity and do not break.                                                                             \\ \midrule
Occlusion         & The video shows a \{view\} view of a \{appearance\}\{object\} smoothly \{object\_movement\} and passing behind a \{occluding\_object\} on a \{surface\} in a \{scene\}. The \{object\} becomes temporarily occluded by the \{occluding\_object\} before reappearing. The \{occluding\_object\} remains stationary throughout the sequence. \\ \midrule
Additional prompt & Everything in the video follows the natural behavior of solid objects in a physical environment. Objects do not fly, morph, or disappear. No new elements appear in the scene. The background is static. Fixed camera view, no camera movement.                                                                                           \\ \bottomrule
\end{tabular}
\label{tab:prompts}
\end{table}

\section{Evaluation Details}
\label{sec:supplementary:evaluation}

\paragraph{Robust minimum and maximum estimation.}

Observing generated videos, we realized that alterations or hallucinations in a few frames can completely change the way humans perceive the video. Therefore, aggregating per-frame results using the average is not appropriate, as the effect of a brief but meaningful error gets diminished. On the other hand, considering only the worst frame is unstable and can be affected by limitations of our metrics. Instead, we compute the average of the worst $k$ frames in the video, an aggregation we denote $\text{RobustMin}$ (or $\text{RobustMax}$, for metrics where larger values are worse) in the definitions below. We select $k$ to be around $5\%$ of the total generated frames. In cases where we compute scores per object, we select the worst $k$ scores across both sets of values. This way, videos where a single object behaves unrealistically do not get their scores averaged out by the other object.

\paragraph{Object segmentation masks and point tracks.}

For all videos, we estimate per-object segmentation masks using SAM3~\citep{carion2025sam} and point tracks using CoTracker3~\citep{karaev2025cotracker3}. In both cases, we prompt the model with the original rendered masks in the first frame, which always matches the ground truth renders. We then propagate through the video to obtain per-frame predictions.

\paragraph{Object disappearance detection.}

Videos in which objects vanish abruptly can yield artificially inflated scores on object-centric metrics, since evaluation is only meaningful on frames where the target object is present. At the same time, given our video settings, we must account for legitimate effects such as occlusion and objects leaving the scene. To disentangle these cases, we introduce an object disappearance detector. A disappearance is registered only when two criteria are jointly satisfied: (i) the object is absent from the disappearance frame onward, and (ii) the object does not exit the image boundary at the disappearance time. We assess the first criterion using the predicted segmentation masks and the second using object tracks, which extrapolate motion beyond the visible frame. Because very rapidly vanishing objects might artificially obtain high scores in some metrics, we conservatively assign videos containing disappearing objects the minimum score on appearance stability, 3D shape stability, and motion similarity.

\paragraph{Occlusion filter.}

Beyond disappearance detection, object-level metrics (appearance stability and 3D shape stability) operate on segmentation masks and therefore require a sufficient number of visible pixels to be reliable. Under severe occlusion, these metrics become unstable. To mitigate this, we introduce an occlusion filter that restricts metric computation to frames with low occlusion. We estimate the occlusion level of an object as the ratio between its current mask size and its mask size in the first frame. We find this simple heuristic to be effective in practice, as the apparent size of objects remains approximately constant across most videos in our setting. For each object, we retain only frames in which the estimated visible fraction exceeds a threshold of $25\%$ of the initial mask size, and we apply this selection independently per object.

\paragraph{Background stability.}

We quantify alterations on the background, which should remain static in all sequences. This way, we are able to detect morphing, new objects appearing, and camera movement. We measure changes by comparing all generated frames to the initial reference frame via pixel-wise MSE:

\begin{equation}
    \text{BackgroundStability} = \text{RobustMax}_t\left(\text{MSE}\left(\text{BG}(I^t), \text{BG}(I^0)\right) \right).
\end{equation}

For every frame, only background pixels shared between both images are considered. The shared background is computed as the intersection between the background mask of both frames.
In this case, we use the robust maximum operation to focus on frames where the background is heavily altered. In order to scale this metric into the $[0,1]$ range, we apply decaying exponential scaling $S'=\exp[-50\cdot S]$.

\paragraph{Motion similarity.}

The introduced metric to evaluate motion is based on the motion encoder from DisMo~\citep{ressler2025dismo}. This model encodes motion representations that are independent from appearance or object identity, while retaining the surrounding scene context of the queried region. This way, we are able to disentangle motion from visual properties, without relying on simplistic metrics, such as comparing object centroid position. We compute the metric by taking the robust minimum over the video of the cosine similarity of the embeddings from the generated video and each different rollout, then select the best match across rollouts,

\begin{equation}
    \text{MotionSimilarity} = \max_r\left(\text{RobustMin}_t\left(\langle\text{Dismo}(I^t)\cdot\text{Dismo}(\tilde{I}_r^t)\rangle\right)\right),
\end{equation}

where $\tilde{I}_r$ indicates the different $r$ reference frames. Before the embeddings are computed, the sequences are temporally aligned. 

\paragraph{Appearance stability.} This metric quantifies semantic distortions of objects across the generated videos. For that, we compare all generated frames with the first context frame, which matches the original render. For a given frame $I^t$, we mask each object ($I_i^t$) and extract its CLS token using DINOv2. Then, we compute the cosine similarity between the tokens from the first and subsequent frames. The stability metric is computed by taking our robust minimum across the video and across objects. This way, the metric is sensitive to object alterations which happen only in a few frames or only happen to a single object. We compute

\begin{equation}
    \text{AppearanceStability} = \text{RobustMin}_{i,t} \left(\langle\text{CLS}(I^t_i)\cdot\text{CLS}(I_i^0)\rangle \right),
\end{equation}

where $\langle \cdot \rangle$ denotes the cosine similarity.

\paragraph{3D shape stability.}

Object shape is usually overlooked or indirectly measured in other benchmarks. We take advantage of advances in 3D shape estimation models to design a novel metric to quantify object morphing in video models. We run SAM3D on the generated videos by prompting the model with the object segmentation masks predicted by SAM3, and obtain object meshes $M_i^t$. For each video, we align all meshes to the reference by optimizing scale and rotation. For each object $i$, we compute the Chamfer Distance (CD) between predicted meshes for the initial and all subsequent frames,

\begin{equation}
    \text{3DShapeStability} = \text{RobustMax}_{i,t} \left(\text{CD}(M_i^t, M_i^0)\right).
\end{equation}

We again apply the exponential scaling used for background stability to bring CD into the $[0,1]$ range.

\paragraph{Physical plausibility.}

This metric captures high-level information about the physical events in the video, covering failure modes that are hard to detect automatically with the low-level metrics above. We design $5$ general questions, as well as $5$ templates for each physical event. The templates are individually adjusted for each video by substituting sequence-specific details, such as object classes. The prompt is:

\begin{mdframed}[backgroundcolor=white!10, linecolor=gray!60!black, roundcorner=8pt, leftmargin=0pt, rightmargin=0pt, innerleftmargin=6pt, innerrightmargin=6pt]
\textbf{VLM-as-judge input prompt} \\
You are evaluating a short video of a physics simulation. An object in the scene was pushed just before the video starts. Watch the video carefully and answer all of the following questions about what you observe. For each question provide a True or False answer. Provide brief comments justifying your answers and rate your overall confidence from 0.0 (not confident at all) to 1.0 (fully confident).

Video description: <video\_description>

Questions: <video\_questions>

Respond ONLY with valid JSON using exactly this structure (no extra text outside the JSON):

\texttt{\{\\
  "answers": \{\\
  \null\quad "question\_key1": <answer1>, \\
  \null\quad ...\\
  \},\\
  "comments": "<brief justification for each answer>", \\
  "confidence": <float 0.0--1.0>\\
\}}

\end{mdframed}

\begin{table}[htbp]
\centering
\caption{\textbf{Questions for physical plausibility metric.}}
\begin{tabular}{lp{10cm}}
\toprule
\textbf{Event}             & \textbf{Questions}\\ \midrule
Fall              &    Does the \{object\_name\} fall off the \{surface\} when it reaches the edge? \\
         &   Does the \{object\_name\} hit the ground? \\
         &   Does the \{object\_name\} change its direction while falling? \\
         &   Does the \{object\_name\} move on an arc-shaped path while falling? \\
         &   Does the \{object\_name\} accelerate while falling?    \\ \midrule
Collision        &       Does the \{object\_name\} contact with the \{collision\_target\}? \\
            & Does the \{object\_name\} come to a stop before contacting the \{collision\_target\}? \\
            & Is the motion of \{object\_name\} affected by the collision with the \{collision\_target\}? \\
            & Is the reaction to the impact realistic considering the size and mass of the objects? \\
            & Are objects deformed or broken during or after the collision? \\ \midrule
Occlusion         &  Does the \{object\_name\} move behind the \{occluding\_object\} during the video? \\
            & Does the \{object\_name\} reappear on the other side after being occluded by the \{occluding\_object\}? \\
            & Does the \{object\_name\} move on a straight path during the whole video? \\
            & Does the \{object\_name\} disappear after being occluded by the \{occluding\_object\}? \\
            & Does the \{object\_name\} change its appearance after being occluded by the \{occluding\_object\}? \\ \midrule
Shared questions & Is the background static throughout the video? \\
                   & Does the \{object\_name\} maintain its color and shape throughout the video? \\
                   & Do new objects appear on the scene during the video? \\
                   & Do objects move smoothly without sudden jumps or teleportation during the video? \\
                   & Do objects move without forces acting on them? \\ \bottomrule
\end{tabular}
\label{tab:vlm-questions}
\end{table}

And the specific questions can be found in \cref{tab:vlm-questions}. The final metric is computed by counting the questions whose answer does not match the expected outcome, and scaling with the inverse of that count:

\begin{equation}
    \text{PhysicalPlausibility} = \left(1 + \sum_{b} \mathbf{1}[b_\text{VLM} \neq b_\text{GT}] \right)^{-1}.
\end{equation}

This way, even a few negative answers result in low physical plausibility, whereas a linear metric would still assign such a video a high score.

\paragraph{Success rate.}

We compress the quality of each video by calibrating thresholds in all our metrics. This way, we prevent catastrophic failures in one metric from being smoothed by other results. The thresholds are calibrated using the human study. We consider successful videos in an annotated metric as those which achieve a median score of $3$ or higher. We then calibrate thresholds on the corresponding metrics by minimizing the absolute difference between false positive rate and false negative rate. 

We study the stability of the thresholds by sweeping them individually and computing the interval of thresholds where the success rate ranking is preserved (Table~\ref{tab:threshold_stats}). The wide ranges show the stability of the selected thresholds. We also bootstrap the subset of 480 human-study videos 1,000 times and use the resamples to compute the confidence interval (CI) of each threshold (Table~\ref{tab:threshold_stats}), which always lies within the ranking-stable interval. We therefore conclude that the thresholds are stable and the rankings are not sensitive to noise in the human study.

\begin{table}[t]
    \centering
    \caption{Statistical study of success thresholds for all parameters. ($^{*}$) Since the physical plausibility metric is discrete, the whole range between $1/3$ and $1/2$ results in the same outcome.}
    \label{tab:threshold_stats}
    \begin{tabular}{lccc}
        \toprule
        Metric & Calibrated threshold & Ranking-stable interval & Calibration 95\% CI \\
        \midrule
        Background Stability       & 0.30 & $[0.08, 0.42]$ & $[0.21, 0.34]$ \\
        Appearance Stability      & 0.48 & $[0.16, 0.62]$ & $[0.45, 0.51]$ \\
        3D Shape Stability   & 0.60 & $[0.00, 0.74]$ & $[0.55, 0.64]$ \\
        Physical Plausibility & 0.49 & $[0.21, 0.50]$ & $[0.33, 0.50]^{*}$ \\
        Motion Similarity     & 0.57 & $[0.55, 0.76]$ & $[0.54, 0.59]$ \\
        \bottomrule
    \end{tabular}
\end{table}

\paragraph{Sensitivity to interventions.}

For each intervention type, we group videos that differ only along that intervention and were generated with the same seed. For a group $g$ and metric $m$, the raw sensitivity $d_m(g)$ is the mean absolute pairwise difference of $m$ between the videos in the group. To make metrics comparable, we normalize by the seed noise $\sigma_m$, the mean absolute pairwise difference of $m$ between seeds of the same input, averaged over all sequences and models. The sensitivity of a model to an intervention is
\begin{equation}
    \text{Sensitivity} = \frac{1}{|M|} \sum_{m \in M} \frac{1}{|G|} \sum_{g \in G} \frac{d_m(g)}{\sigma_m},
\end{equation}
where $M$ is the set of metrics and $G$ the set of groups for that intervention across all seeds. The average sensitivity in \cref{tab:sensitivities_per_event} is the mean over the interventions available for each event.

Since $\sigma_m$ is averaged over all models, it is a single constant per metric that is shared by every model. It therefore only rescales each metric and does not depend on a model's own seed variability. A sensitivity of $1$ means that an intervention changes quality as much as resampling the seed typically does.

Note that sensitivity measures the consistency of generation \emph{quality}, not the similarity between generated videos. Since each video is evaluated against the references of its own input, two videos with different trajectories, e.g., after an object intervention changes the object's mass, obtain identical scores if both match their respective references.

\clearpage

\section{Human Study}\label{sec:supplementary:userstudy}

\begin{figure}[t]
    \centering
    \includegraphics[width=0.8\linewidth]{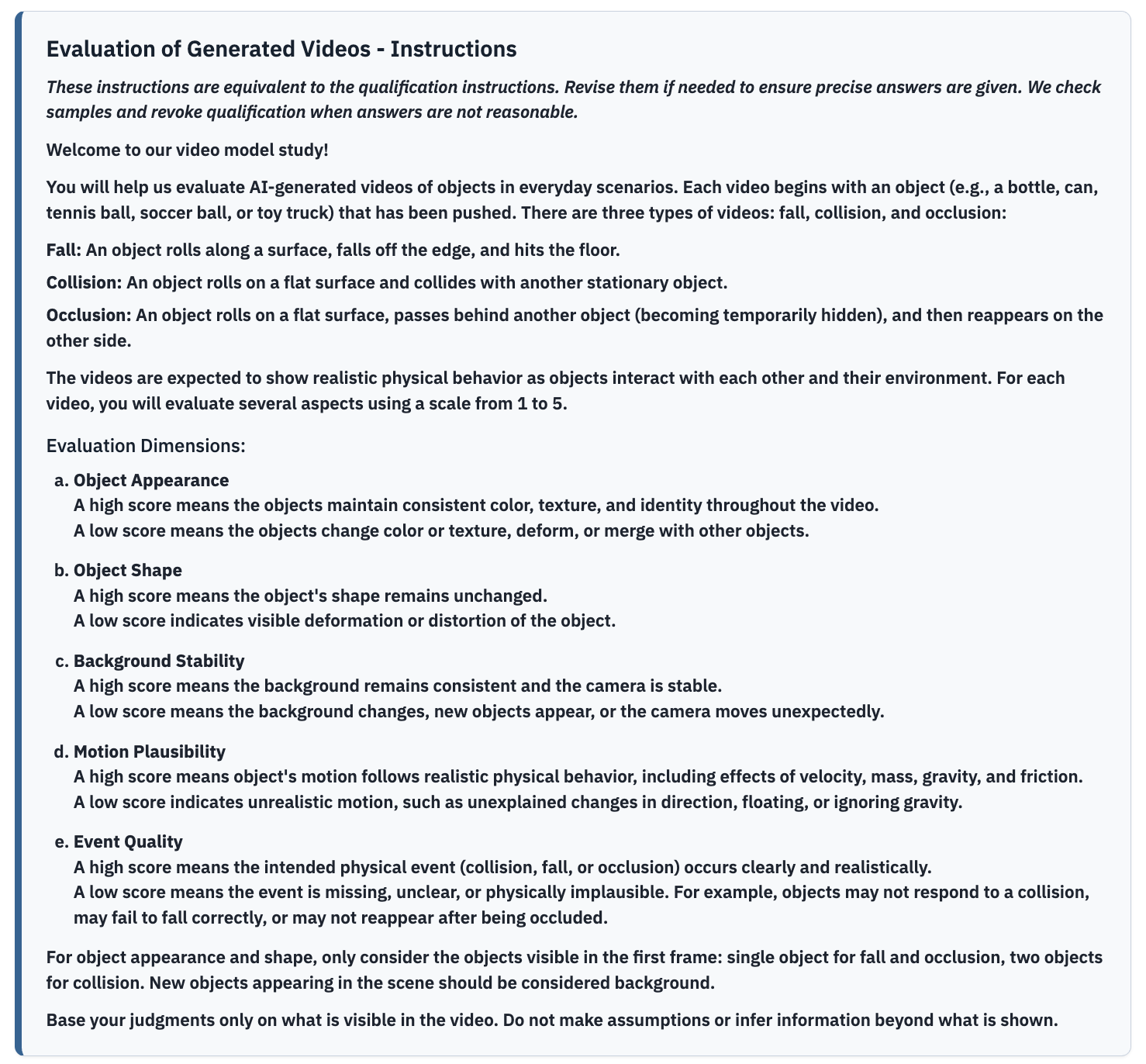}
    \caption{Instructions of the human study.}
    \label{fig:userstudy-instructions}
\end{figure}

\begin{figure}[t]
    \centering
    \includegraphics[width=0.7\linewidth]{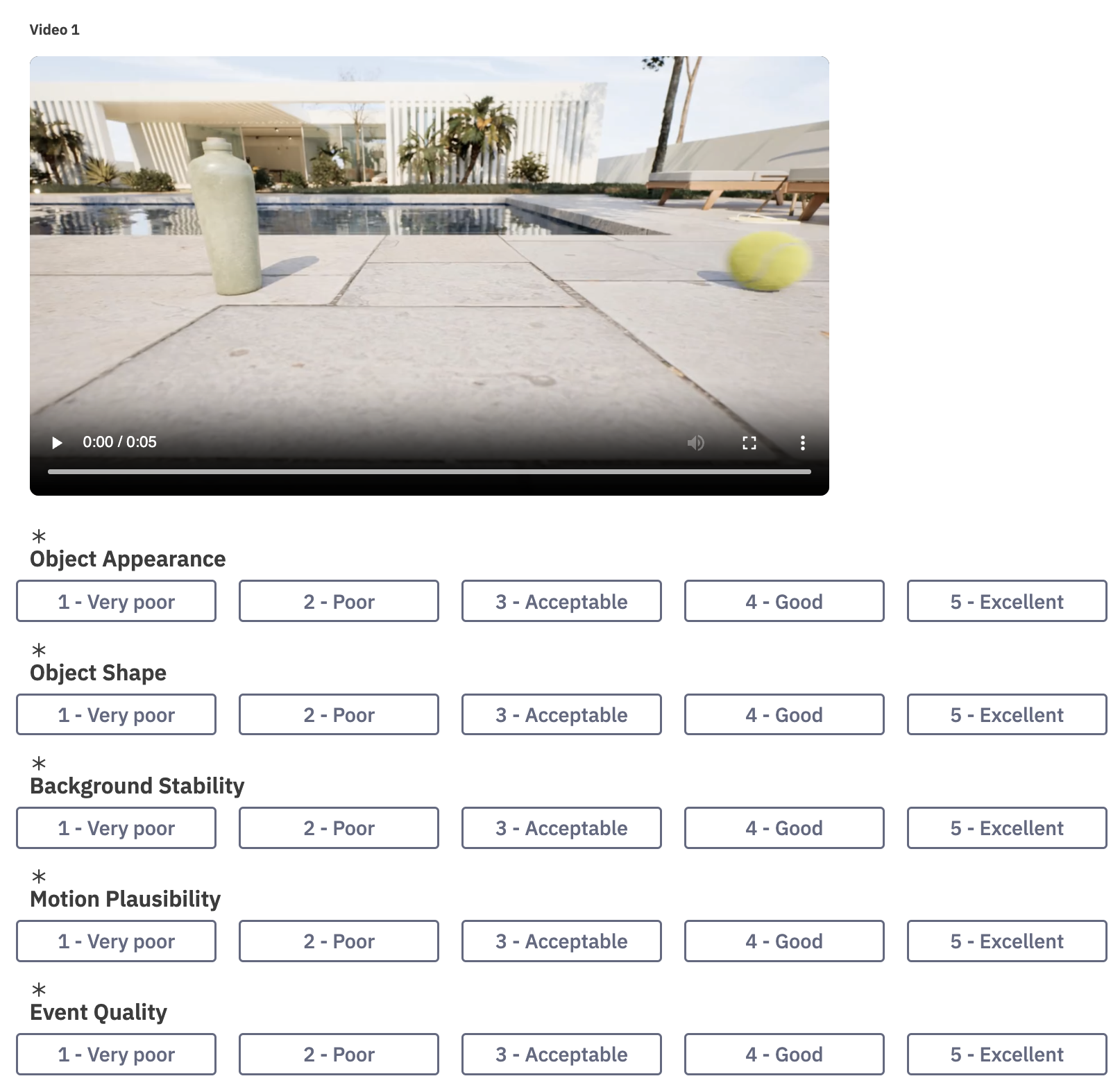}
    \caption{GUI used in the user study for annotating video quality.}
    \label{fig:userstudy-example-annotation}
\end{figure}

We present the instructions of the human study in \cref{fig:userstudy-instructions} and the GUI for annotating videos along the considered quality axes in \cref{fig:userstudy-example-annotation}.
All annotators received detailed instructions and needed to pass a qualification exam before participation.
The $8$ hired Prolific annotators received a compensation of \pounds $14$ per hour.

\clearpage

\section{Additional Examples}

\label{supp:additional_plots}
\begin{figure}[htbp]
    \centering
    \includegraphics[width=1\linewidth]{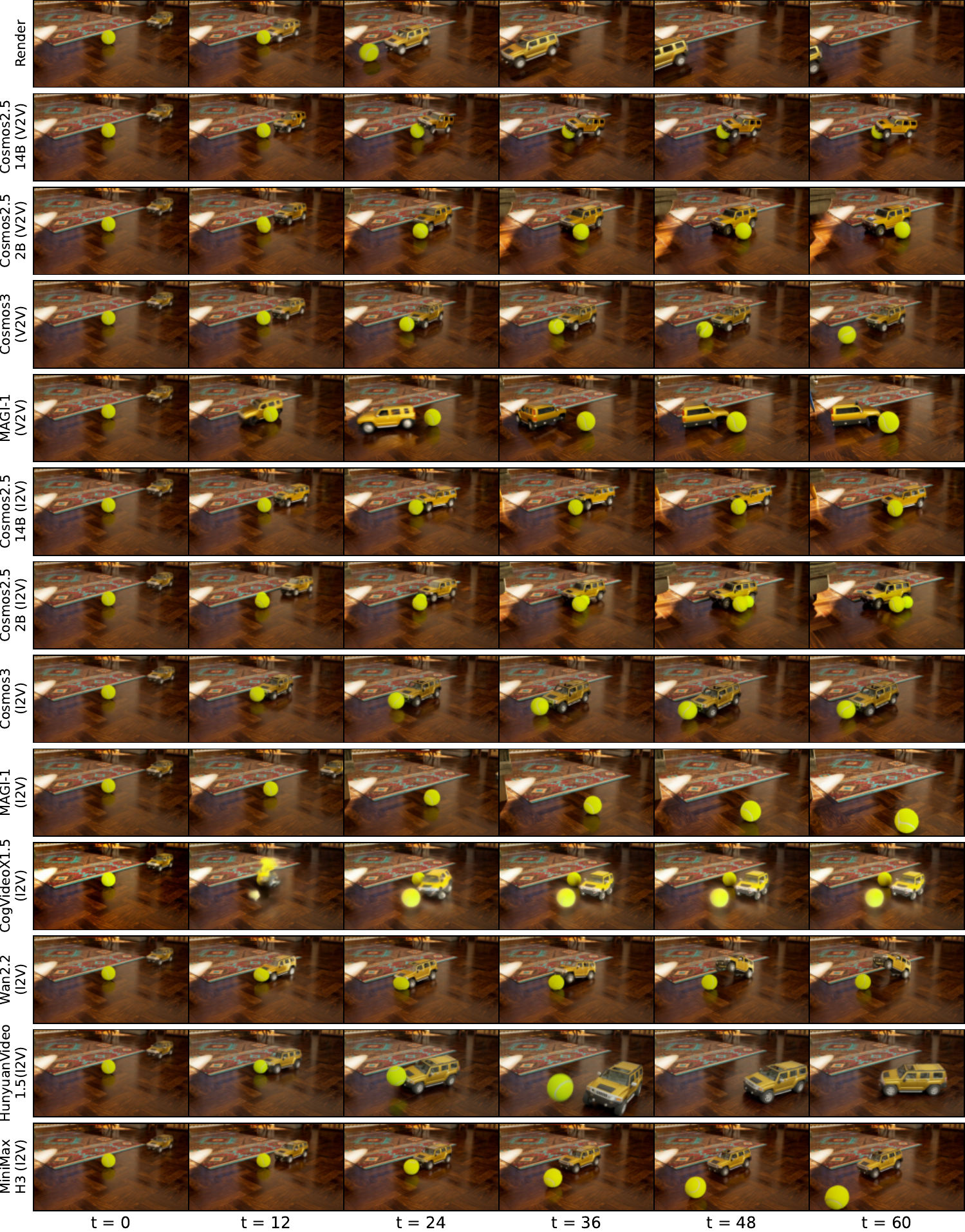}
    \caption{\textbf{Additional generated examples: collision.}}
    \label{fig:examples1}
\end{figure}

\begin{figure}[htbp]
    \centering
    \includegraphics[width=1\linewidth]{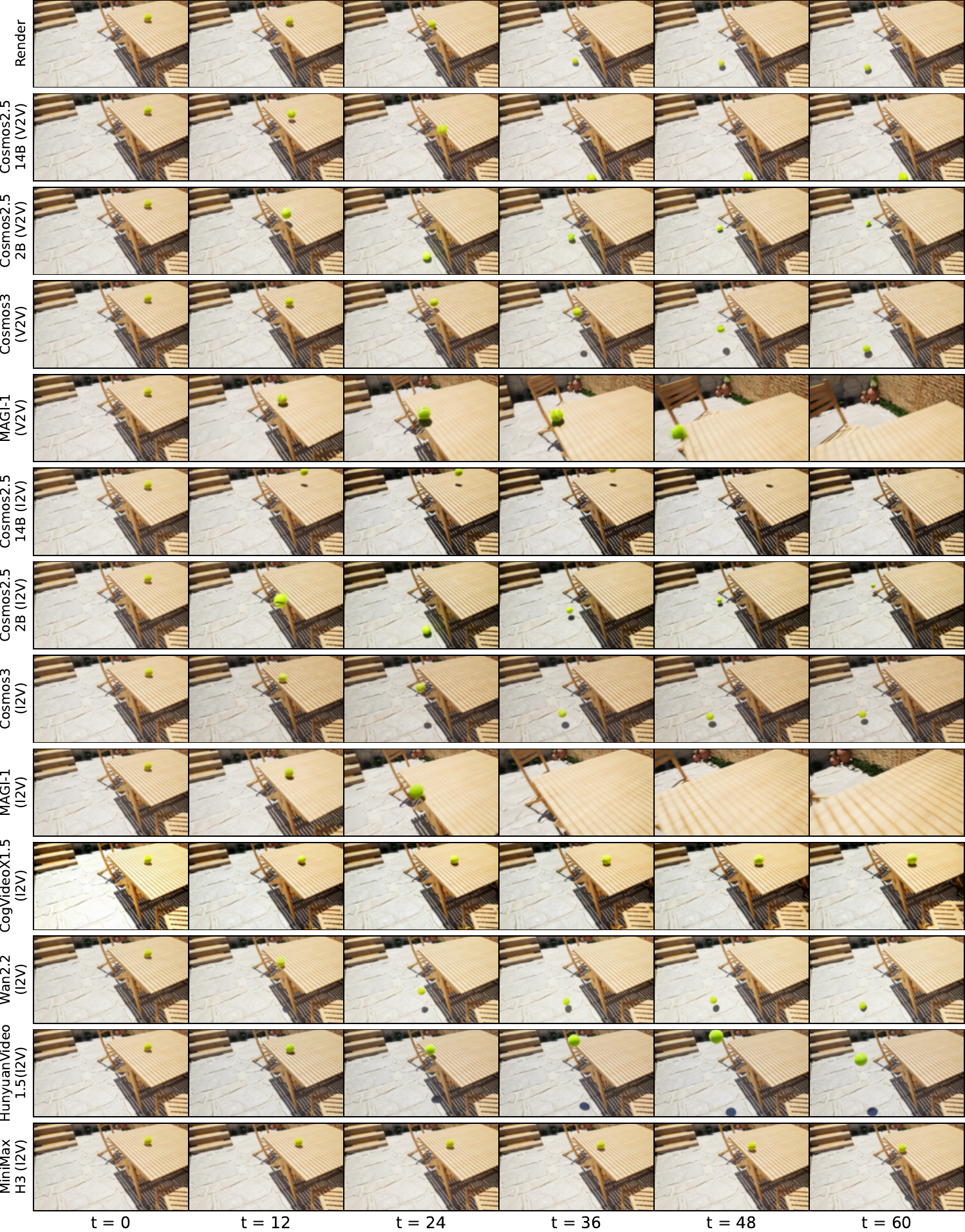}
    \caption{\textbf{Additional generated examples: fall.}}
    \label{fig:examples2}
\end{figure}

\begin{figure}[htbp]
    \centering
    \includegraphics[width=1\linewidth]{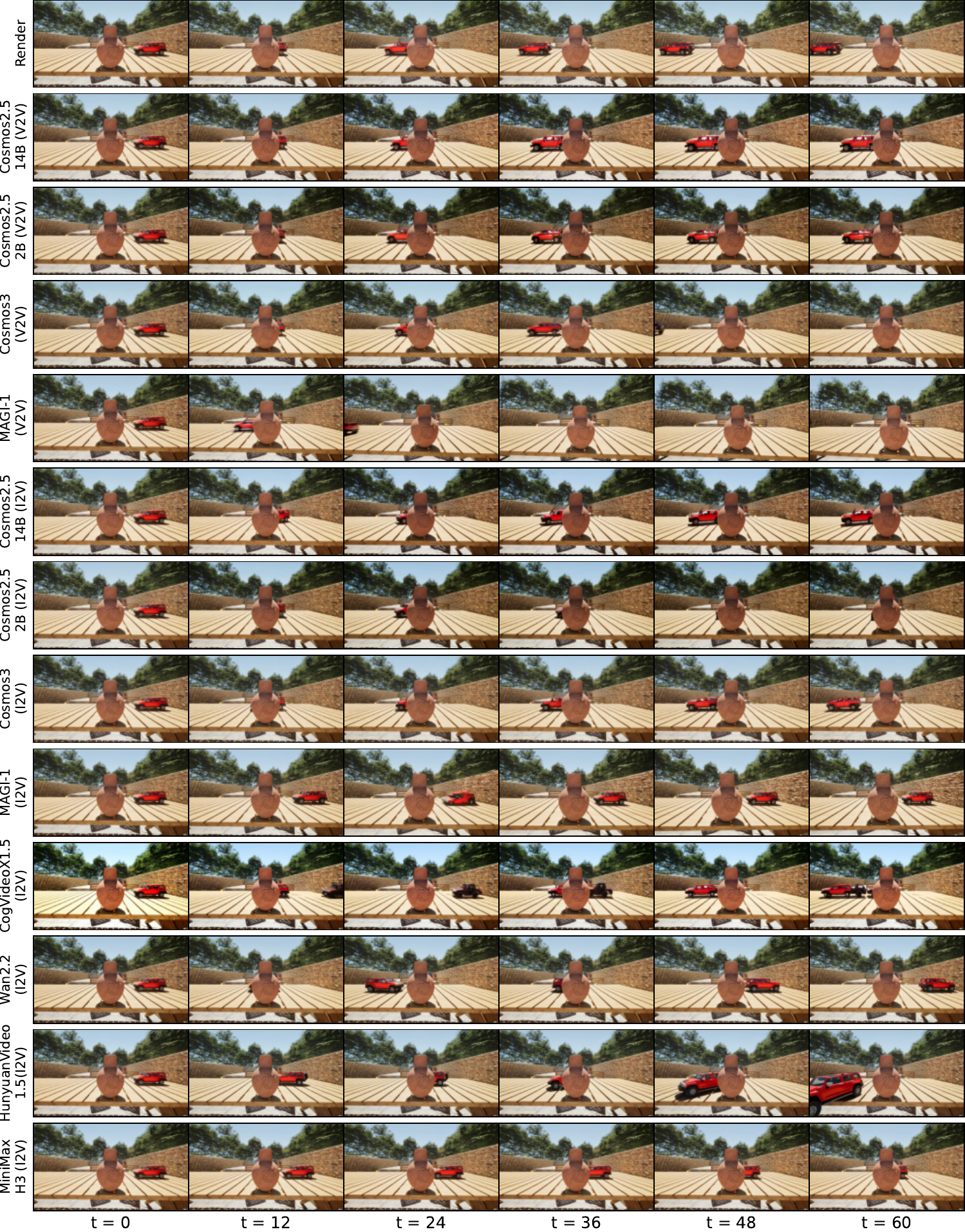}
    \caption{\textbf{Additional generated examples: occlusion.}}
    \label{fig:examples3}
\end{figure}

In \cref{fig:examples1,fig:examples2,fig:examples3} we show rollout examples of generated videos.

\clearpage

\section{Real Data Experiment}
\label{supp:real_data_experiment}

We conduct a small-scale experiment to estimate the simulation-to-real gap on our benchmark. We imitate benchmark sequences by recording experiments in a real environment. In particular, we run all three physical events, using three different objects. After that, we generate videos for all models following the benchmark's prompt structure in the I2V setting, for a total of $315$ generated videos.

\begin{figure}
    \centering
    \includegraphics[width=0.9\linewidth]{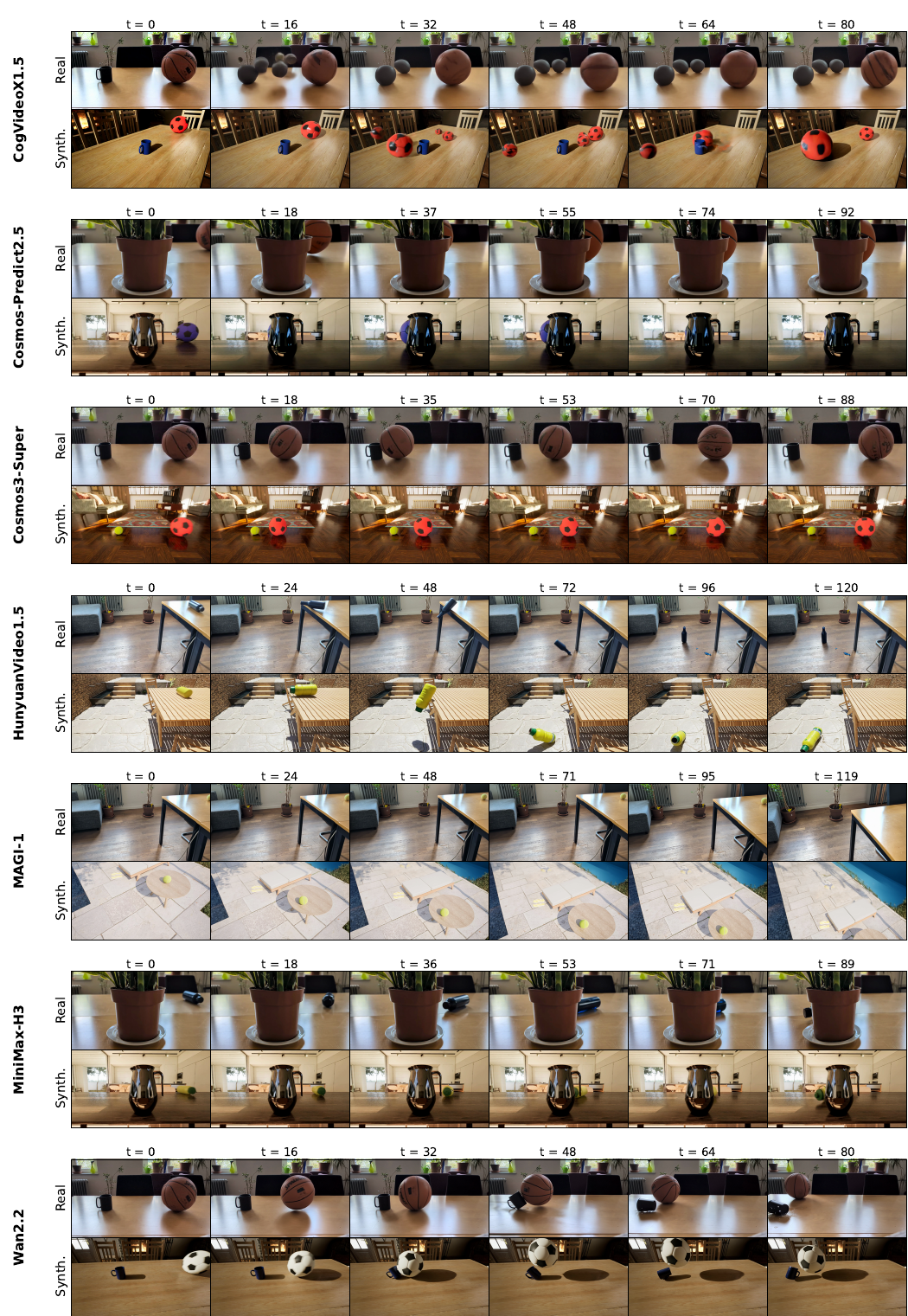}
    \caption{\textbf{Real and synthetic data rollout comparison.} Each model shows a different common failure mode of the models. In order: severe object distortion and duplication, unexplained stop, unexplained back motion, object distorting and reversing sides, undesired camera motion, unexplained rotation and incorrect physical properties. The fact that similar failure modes are observed in both data regimes suggest a small gap in the abilities the benchmark quantifies.}
\label{fig:real_data_experiment}
\end{figure}

The numerical results (\cref{tab:real_metrics_global}) show similar results as for the main benchmark, with MiniMax-H3 and Cosmos3-Super achieving the highest success rates and MAGI-1 and CogVideoX1.5 performing worst. In general, all metrics stay in similar ranges as for synthetic data. The qualitative results in \cref{fig:real_data_experiment} also show similar failure modes in both data regimes. These experiments suggest that the simulation-to-real gap is small.

\begin{table}[ht]
\centering
\caption{\textbf{Benchmark metrics on real videos.}}
\label{tab:real_metrics_global}
\small
\begin{tabular}{cccccccc}
\toprule
Mode & Model & \makecell{Bg \\ Stab.} & \makecell{Motion \\ Sim.} & \makecell{App. \\ Stab.} & \makecell{3D Shape\\ Stab.} & \makecell{Physical \\ Plaus.} & \makecell{Success \\ Rate} \\
\midrule
\multirow{7}{*}{I2V} & Cosmos2.5-2B & 0.44 & 0.42 & 0.54 & 0.74 & 0.52 & 0.04 \\
 & Cosmos3-Super & \underline{0.64} & \underline{0.60} & 0.58 & 0.76 & \textbf{0.61} & \underline{0.09} \\
 & Magi-1-4.5B & 0.25 & 0.49 & \underline{0.65} & \underline{0.81} & 0.41 & 0.00 \\
 & CogVideoX1.5-5B & 0.21 & 0.37 & 0.57 & 0.70 & 0.43 & 0.00 \\
 & Wan2.2-14B & \dashuline{0.58} & \dashuline{0.56} & \dashuline{0.60} & \dashuline{0.79} & \underline{0.60} & \dashuline{0.07} \\
 & HunyuanVideo-1.5 & 0.51 & 0.46 & 0.56 & 0.70 & 0.55 & 0.02 \\
 & MiniMax H3 & \textbf{0.73} & \textbf{0.61} & \textbf{0.67} & \textbf{0.86} & \dashuline{0.58} & \textbf{0.16} \\
\bottomrule
\end{tabular}
\end{table}

\clearpage

\section{Detailed Results}
\label{sec:supplementary:results}

\begin{figure}[htbp]
    \centering
    \includegraphics[width=0.75\linewidth]{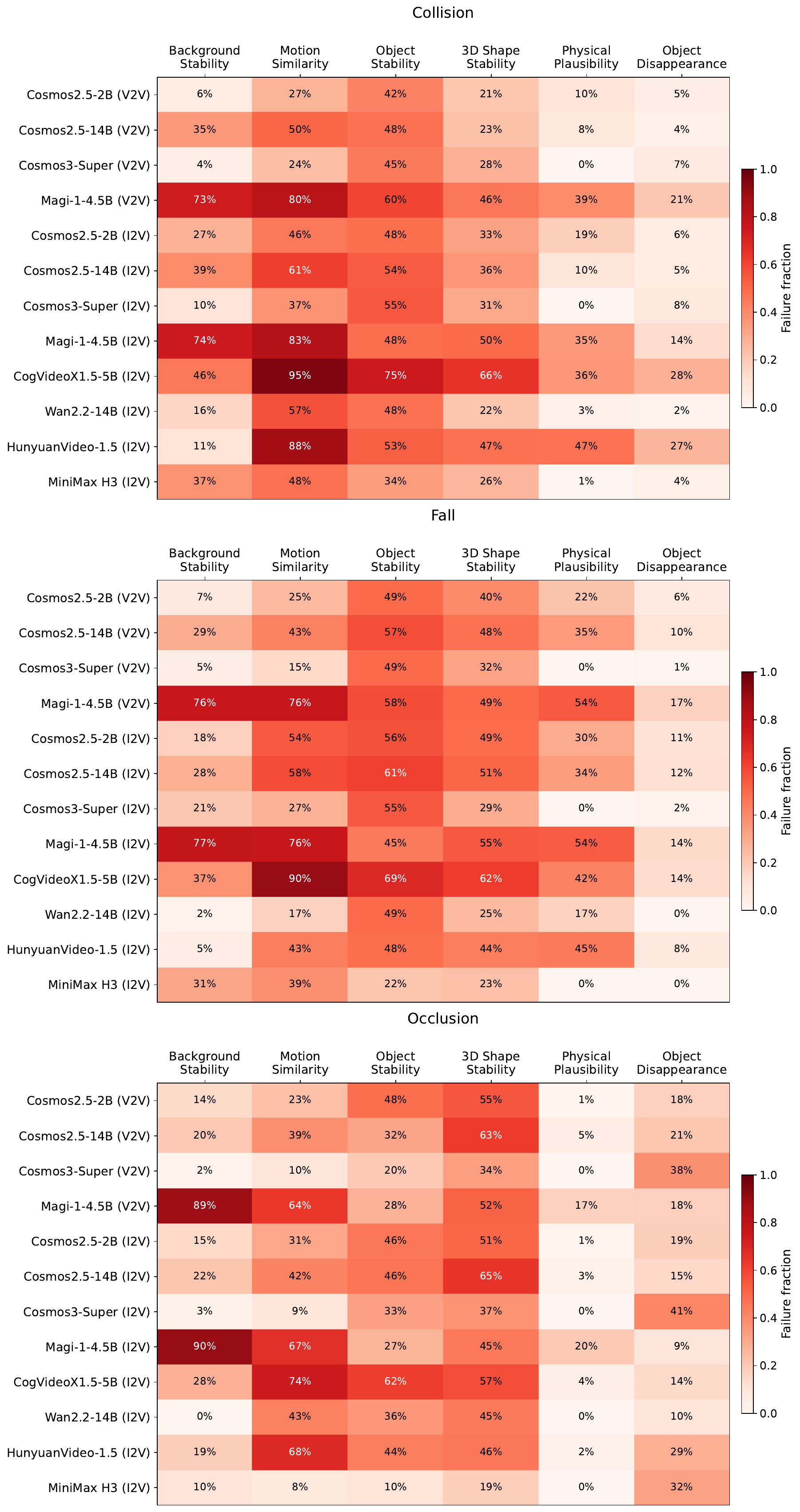}
    \caption{\textbf{Failure rates per metric.} For each event, we provide the fraction of videos failing in each metric.}
    \label{fig:failure_rates_per_event}
\end{figure}

We provide benchmark metrics for each physical event in \cref{tab:metrics_per_event} and sensitivity to interventions in \cref{tab:sensitivities_per_event}. We also provide the failure rates per metric in \cref{fig:failure_rates_per_event}.

\begin{table}[htbp]
\centering
\caption{\textbf{Benchmark metrics per physical event.} Metrics are normalized to $[0,1]$ and higher is better. Best, second-, and third-best values per metric are indicated by bold, underline, and dashed underline, respectively. Values are ranked within each event, across both conditioning modes.}
\label{tab:metrics_per_event}
\small
\begin{tabular}{cccccccc}
\toprule
Mode & Model & \makecell{Bg \\ Stab.} & \makecell{Motion \\ Sim.} & \makecell{App. \\ Stab.} & \makecell{3D Shape\\ Stab.} & \makecell{Physical \\ Plaus.} & \makecell{Success \\ Rate} \\
\midrule
\multicolumn{8}{l}{\textit{Fall}} \\
\midrule
\multirow{4}{*}{V2V} & Cosmos2.5-2B & \dashuline{0.80} & \dashuline{0.64} & 0.52 & 0.62 & 0.47 & 0.16 \\
 & Cosmos2.5-14B & 0.59 & 0.55 & 0.49 & 0.56 & 0.43 & 0.10 \\
 & Cosmos3-Super & \underline{0.82} & \underline{0.67} & 0.54 & 0.68 & \underline{0.87} & \underline{0.27} \\
 & Magi-1-4.5B & 0.23 & 0.48 & 0.47 & 0.55 & 0.38 & 0.00 \\
\midrule
\multirow{8}{*}{I2V} & Cosmos2.5-2B & 0.64 & 0.54 & 0.48 & 0.56 & 0.45 & 0.08 \\
 & Cosmos2.5-14B & 0.57 & 0.50 & 0.47 & 0.55 & 0.43 & 0.06 \\
 & Cosmos3-Super & 0.67 & 0.63 & 0.52 & \dashuline{0.70} & \textbf{0.88} & 0.19 \\
 & Magi-1-4.5B & 0.22 & 0.47 & 0.52 & 0.55 & 0.39 & 0.02 \\
 & CogVideoX1.5-5B & 0.41 & 0.37 & 0.42 & 0.49 & 0.41 & 0.01 \\
 & Wan2.2-14B & \textbf{0.83} & \textbf{0.68} & \underline{0.56} & \underline{0.73} & 0.47 & \dashuline{0.24} \\
 & HunyuanVideo-1.5 & 0.76 & 0.58 & \dashuline{0.55} & 0.62 & 0.42 & 0.09 \\
 & MiniMax H3 & 0.62 & 0.53 & \textbf{0.66} & \textbf{0.75} & \dashuline{0.84} & \textbf{0.40} \\
\midrule
\multicolumn{8}{l}{\textit{Collision}} \\
\midrule
\multirow{4}{*}{V2V} & Cosmos2.5-2B & \textbf{0.79} & \underline{0.65} & \underline{0.54} & \textbf{0.77} & \dashuline{0.90} & \textbf{0.31} \\
 & Cosmos2.5-14B & 0.51 & 0.56 & \dashuline{0.52} & \underline{0.76} & 0.86 & \dashuline{0.18} \\
 & Cosmos3-Super & \underline{0.78} & \textbf{0.67} & \dashuline{0.52} & 0.73 & \textbf{0.94} & \textbf{0.31} \\
 & Magi-1-4.5B & 0.24 & 0.45 & 0.45 & 0.61 & 0.59 & 0.02 \\
\midrule
\multirow{8}{*}{I2V} & Cosmos2.5-2B & 0.57 & 0.58 & 0.51 & 0.70 & 0.80 & 0.17 \\
 & Cosmos2.5-14B & 0.46 & 0.52 & 0.50 & 0.69 & 0.85 & 0.10 \\
 & Cosmos3-Super & \dashuline{0.72} & \dashuline{0.61} & 0.48 & 0.70 & \underline{0.93} & 0.16 \\
 & Magi-1-4.5B & 0.24 & 0.44 & 0.49 & 0.57 & 0.65 & 0.02 \\
 & CogVideoX1.5-5B & 0.36 & 0.30 & 0.38 & 0.47 & 0.65 & 0.01 \\
 & Wan2.2-14B & 0.67 & 0.53 & 0.50 & \dashuline{0.75} & \textbf{0.94} & 0.17 \\
 & HunyuanVideo-1.5 & 0.66 & 0.42 & 0.48 & 0.59 & 0.42 & 0.02 \\
 & MiniMax H3 & 0.54 & 0.53 & \textbf{0.57} & 0.73 & \textbf{0.94} & \underline{0.28} \\
\midrule
\multicolumn{8}{l}{\textit{Occlusion}} \\
\midrule
\multirow{4}{*}{V2V} & Cosmos2.5-2B & 0.59 & 0.64 & 0.52 & 0.54 & 0.93 & 0.19 \\
 & Cosmos2.5-14B & 0.55 & 0.60 & 0.56 & 0.50 & 0.88 & 0.14 \\
 & Cosmos3-Super & 0.77 & \textbf{0.73} & \dashuline{0.58} & \underline{0.72} & \textbf{1.00} & \underline{0.35} \\
 & Magi-1-4.5B & 0.11 & 0.52 & 0.55 & 0.59 & 0.74 & 0.02 \\
\midrule
\multirow{8}{*}{I2V} & Cosmos2.5-2B & 0.59 & 0.61 & 0.52 & 0.61 & 0.92 & 0.15 \\
 & Cosmos2.5-14B & 0.56 & 0.58 & 0.51 & 0.51 & 0.89 & 0.14 \\
 & Cosmos3-Super & \dashuline{0.82} & \dashuline{0.69} & 0.54 & \dashuline{0.70} & \textbf{1.00} & \dashuline{0.31} \\
 & Magi-1-4.5B & 0.10 & 0.50 & \underline{0.59} & 0.62 & 0.74 & 0.05 \\
 & CogVideoX1.5-5B & 0.43 & 0.48 & 0.44 & 0.55 & 0.93 & 0.08 \\
 & Wan2.2-14B & \underline{0.84} & 0.58 & \dashuline{0.58} & 0.63 & \dashuline{0.97} & 0.22 \\
 & HunyuanVideo-1.5 & 0.55 & 0.49 & 0.54 & 0.60 & 0.95 & 0.12 \\
 & MiniMax H3 & \textbf{0.85} & \underline{0.72} & \textbf{0.69} & \textbf{0.82} & \underline{0.99} & \textbf{0.49} \\
\bottomrule
\end{tabular}
\end{table}

\begin{table}[htbp]
\centering
\caption{\textbf{Benchmark sensitivities per physical event.} Lower is better: a lower value indicates that generation quality changes less under that intervention. Best, second-, and third-best values per metric are indicated by bold, underline, and dashed underline, respectively. Values are ranked within each event, across both conditioning modes. Viewpoint is not varied for occlusion events, so those entries are marked ``--''.}
\label{tab:sensitivities_per_event}
\small
\begin{tabular}{ccccccc}
\toprule
Mode & Model & Scene & Object & Appearance & View & Average \\
\midrule
\multicolumn{7}{l}{\textit{Fall}} \\
\midrule
\multirow{4}{*}{V2V} & Cosmos2.5-2B & 1.14 & 1.17 & 0.73 & \dashuline{1.09} & 1.03 \\
 & Cosmos2.5-14B & 1.55 & 1.27 & 0.75 & 1.28 & 1.21 \\
 & Cosmos3-Super & \underline{0.90} & \textbf{1.08} & \underline{0.53} & \textbf{0.85} & \textbf{0.84} \\
 & Magi-1-4.5B & 1.36 & 1.36 & 1.00 & 1.41 & 1.28 \\
\midrule
\multirow{8}{*}{I2V} & Cosmos2.5-2B & 1.34 & 1.35 & 0.83 & 1.33 & 1.21 \\
 & Cosmos2.5-14B & 1.53 & 1.33 & 0.79 & 1.19 & 1.21 \\
 & Cosmos3-Super & \dashuline{1.06} & \dashuline{1.14} & 0.57 & 1.10 & \dashuline{0.97} \\
 & Magi-1-4.5B & 1.42 & 1.34 & 1.04 & 1.50 & 1.33 \\
 & CogVideoX1.5-5B & 1.25 & 1.32 & 1.04 & 1.32 & 1.23 \\
 & Wan2.2-14B & \textbf{0.89} & 1.18 & \dashuline{0.54} & \underline{0.87} & \underline{0.87} \\
 & HunyuanVideo-1.5 & 1.12 & 1.41 & 0.75 & 1.10 & 1.09 \\
 & MiniMax H3 & 1.47 & \underline{1.09} & \textbf{0.48} & 1.24 & 1.07 \\
\midrule
\multicolumn{7}{l}{\textit{Collision}} \\
\midrule
\multirow{4}{*}{V2V} & Cosmos2.5-2B & \underline{1.18} & 1.19 & 0.68 & \underline{1.12} & \underline{1.04} \\
 & Cosmos2.5-14B & 1.50 & 1.37 & 0.78 & 1.42 & 1.26 \\
 & Cosmos3-Super & \textbf{1.07} & \underline{1.05} & \underline{0.60} & \textbf{1.08} & \textbf{0.95} \\
 & Magi-1-4.5B & 1.62 & 1.46 & 1.06 & 1.65 & 1.45 \\
\midrule
\multirow{8}{*}{I2V} & Cosmos2.5-2B & 1.57 & 1.48 & 0.93 & 1.53 & 1.38 \\
 & Cosmos2.5-14B & 1.55 & 1.45 & 0.89 & 1.42 & 1.33 \\
 & Cosmos3-Super & \dashuline{1.19} & \dashuline{1.18} & \dashuline{0.67} & \dashuline{1.13} & \underline{1.04} \\
 & Magi-1-4.5B & 1.81 & 1.48 & 1.16 & 1.71 & 1.54 \\
 & CogVideoX1.5-5B & 1.59 & 1.52 & 1.25 & 1.53 & 1.47 \\
 & Wan2.2-14B & 1.25 & 1.19 & 0.69 & 1.14 & \dashuline{1.07} \\
 & HunyuanVideo-1.5 & 1.26 & 1.25 & 0.82 & 1.17 & 1.13 \\
 & MiniMax H3 & 1.48 & \textbf{1.04} & \textbf{0.57} & 1.52 & 1.15 \\
\midrule
\multicolumn{7}{l}{\textit{Occlusion}} \\
\midrule
\multirow{4}{*}{V2V} & Cosmos2.5-2B & 1.13 & 1.18 & 0.72 & -- & 1.01 \\
 & Cosmos2.5-14B & 1.35 & 1.18 & 0.77 & -- & 1.10 \\
 & Cosmos3-Super & \dashuline{0.75} & \textbf{0.81} & \underline{0.49} & -- & \underline{0.69} \\
 & Magi-1-4.5B & 1.33 & 1.28 & 1.04 & -- & 1.22 \\
\midrule
\multirow{8}{*}{I2V} & Cosmos2.5-2B & 1.21 & \dashuline{1.09} & 0.77 & -- & 1.02 \\
 & Cosmos2.5-14B & 1.36 & 1.23 & 0.88 & -- & 1.16 \\
 & Cosmos3-Super & \underline{0.70} & \underline{0.88} & \dashuline{0.50} & -- & \underline{0.69} \\
 & Magi-1-4.5B & 1.45 & 1.36 & 1.12 & -- & 1.31 \\
 & CogVideoX1.5-5B & 1.22 & 1.33 & 1.07 & -- & 1.21 \\
 & Wan2.2-14B & 0.82 & 1.11 & 0.61 & -- & \dashuline{0.85} \\
 & HunyuanVideo-1.5 & 1.22 & 1.41 & 0.84 & -- & 1.16 \\
 & MiniMax H3 & \textbf{0.63} & \textbf{0.81} & \textbf{0.31} & -- & \textbf{0.58} \\
\bottomrule
\end{tabular}
\end{table}

\begin{table}[htbp]
\centering
\small
\caption{\textbf{Human study results.} Average median human annotation score per model and metric, $1$--$5$ scale, higher is better. Best, second-, and third-best values per metric are indicated by bold, underline, and dashed underline, respectively.}
\begin{tabular}{ccccccc}
\toprule
Mode & Model & \makecell{Object \\ Appear.} & \makecell{Object \\ Shape} & \makecell{Bg \\ Stab.} & \makecell{Motion \\ Plaus.} & \makecell{Event \\ Quality} \\
\midrule
\multirow{3}{*}{V2V} & Cosmos2.5-2B & \dashuline{3.65} & \dashuline{3.30} & \textbf{4.44} & \dashuline{1.62} & 1.88 \\
 & Cosmos2.5-14B & \underline{3.70} & \underline{3.33} & 3.88 & \underline{1.76} & \underline{1.94} \\
 & MAGI-1 & 2.57 & 2.57 & 2.26 & 1.40 & 1.33 \\
\midrule
\multirow{5}{*}{I2V} & Cosmos2.5-2B & 3.42 & 3.29 & \dashuline{3.92} & 1.53 & 1.68 \\
 & Cosmos2.5-14B & 3.47 & 3.14 & 3.83 & \underline{1.76} & \dashuline{1.91} \\
 & MAGI-1 & 2.62 & 2.70 & 2.20 & 1.38 & 1.29 \\
 & CogVideoX1.5 & 1.59 & 1.37 & 2.78 & 1.08 & 1.16 \\
 & Wan2.2 & \textbf{4.15} & \textbf{4.03} & \underline{4.21} & \textbf{2.33} & \textbf{2.68} \\
\bottomrule
\end{tabular}

\label{tab:human_study}
\end{table}

\clearpage

\section{Asset Licenses and Release Documentation}
\label{sec:supplementary:asset-licenses}

\modelname{} uses third-party Unreal Engine scene assets to construct the rendered simulation environments. All 3D assets were purchased under Epic Content License Agreement (ECLA), and selected only when authors allowed GenAI-related research and benchmark generation. We use the assets as part of controlled synthetic scenes rendered in Unreal Engine; the benchmark does not rely on scraped videos or unlicensed real-world footage.

We release the generated benchmark videos and evaluation code on the project page (\projecturl{}). The release includes the rendered RGB videos, text prompts, benchmark metadata describing the physical event and intervention values for each video, and the code used to compute the evaluation metrics reported in the paper.

We do not redistribute third-party Unreal Engine source assets whose licenses restrict redistribution. Instead, released artifacts are limited to the generated benchmark data, code, and metadata that can be shared under the applicable licenses. The released package includes license information for the dataset and code, along with attribution and license notes for third-party tools and assets used to construct the benchmark.

\end{document}